\documentclass[]{TEAI}

\usepackage{amsmath}
\usepackage{url}  
\usepackage{booktabs}  
\usepackage{multirow} 

\providecommand{\shortcite}{}
\renewcommand{\shortcite}{\citeyearpar}

\usepackage{algorithm}
\usepackage{algorithmic}
\tcbset{
  aibox/.style={
    top=10pt,
    colback=blue!2,
    colframe=blue!40!black!70,
    fontupper=\small,
    colbacktitle=blue!30,
    coltitle=black,
    fonttitle=\bfseries,
    enhanced,
    center,
    breakable,
    attach boxed title to top left={yshift=-0.1in,xshift=0.15in},
    boxed title style={boxrule=0pt,colframe=white,},
    fontupper=\footnotesize,
  }
}
\newtcolorbox{AIbox}[2][]{aibox, title=#2,#1}
\newcommand{\ours}{ScriptMoE}

\title{All-in-One Multilingual Scene Text Recognition with Script-aware Mixture-of-Experts}

\author{
    Xingsong Ye\textsuperscript{1,2},
    Yongkun Du\textsuperscript{1,2},
    Jiaxin Zhang\textsuperscript{3},
    Zhixian Li\textsuperscript{1,2},
    Chong Sun\textsuperscript{3}, \\
    Chen Li\textsuperscript{3}, 
    Jing LYU\textsuperscript{3}, Lianwen Jin\textsuperscript{4}, Zhineng Chen\textsuperscript{1,2,$\dagger$}
}

\affiliation[1]{\mbox{Institute of Trustworthy Embodied AI, Fudan University}}
\affiliation[2]{\mbox{Shanghai Key Laboratory of Multimodal Embodied AI}}
\affiliation[3]{\mbox{WeChat Vision, Tencent Inc.}}
\affiliation[4]{\mbox{South China University of Technology}}

\abstract{
\begin{abstract}

Multilingual scene text recognition (STR) remains challenging due to the scarcity of training data for most languages and the difficulty of serving diverse scripts within a single model. Existing solutions either deploy one recognizer per language, inflating cost and introducing error accumulation, or rely on massive vision-language models (VLMs) that are expensive and still inaccurate on many scripts. In this work, we pursue an all-in-one multilingual recognizer that is simpler than per-language experts, lighter than VLMs, and more accurate than both. First, we construct TextMuSS-10M, a large-scale synthetic scene text dataset spanning 10 scripts and 229 languages. It provides balanced and sufficient supervision where real data is unavailable. Second, we propose ScriptMoE, a script-aware Mixture-of-Experts (MoE) architecture. It shares a single visual encoder and replaces the dense decoder with a sparse MoE block, which consists of an image-level router dispatches each image to the top-2 script-aligned experts and a shared expert absorbs cross-script knowledge. Extensive experiments on our assembled TextMuSS-Bench (10 scripts, 10,899 images) show that ScriptMoE achieves the highest accuracy of 82.06\%, outperforming the strongest STR baseline by 1.31\%. On the CC-OCR end-to-end multilingual task, replacing only the recognizer in PP-OCRv5 with ScriptMoE lifts F1 score from 65.71\% to 80.89\%, slightly surpassing the best VLM (80.73\%) at a fraction of the parameter count.
\end{abstract}
}

\correspondence{\email{zhinchen@fudan.edu.cn}}
\checkdata[Code]{\url{https://github.com/YesianRohn/ScriptMoE} \& \url{https://github.com/Topdu/OpenOCR}}

\begin{document}
\maketitle
\renewcommand{\thefootnote}{}
\footnotetext{$^\dagger$Corresponding authors.}
\renewcommand{\thefootnote}{\arabic{footnote}}

\vspace{-1.5em}

\begin{figure}[t]
\centering
\includegraphics[width=0.8\textwidth]{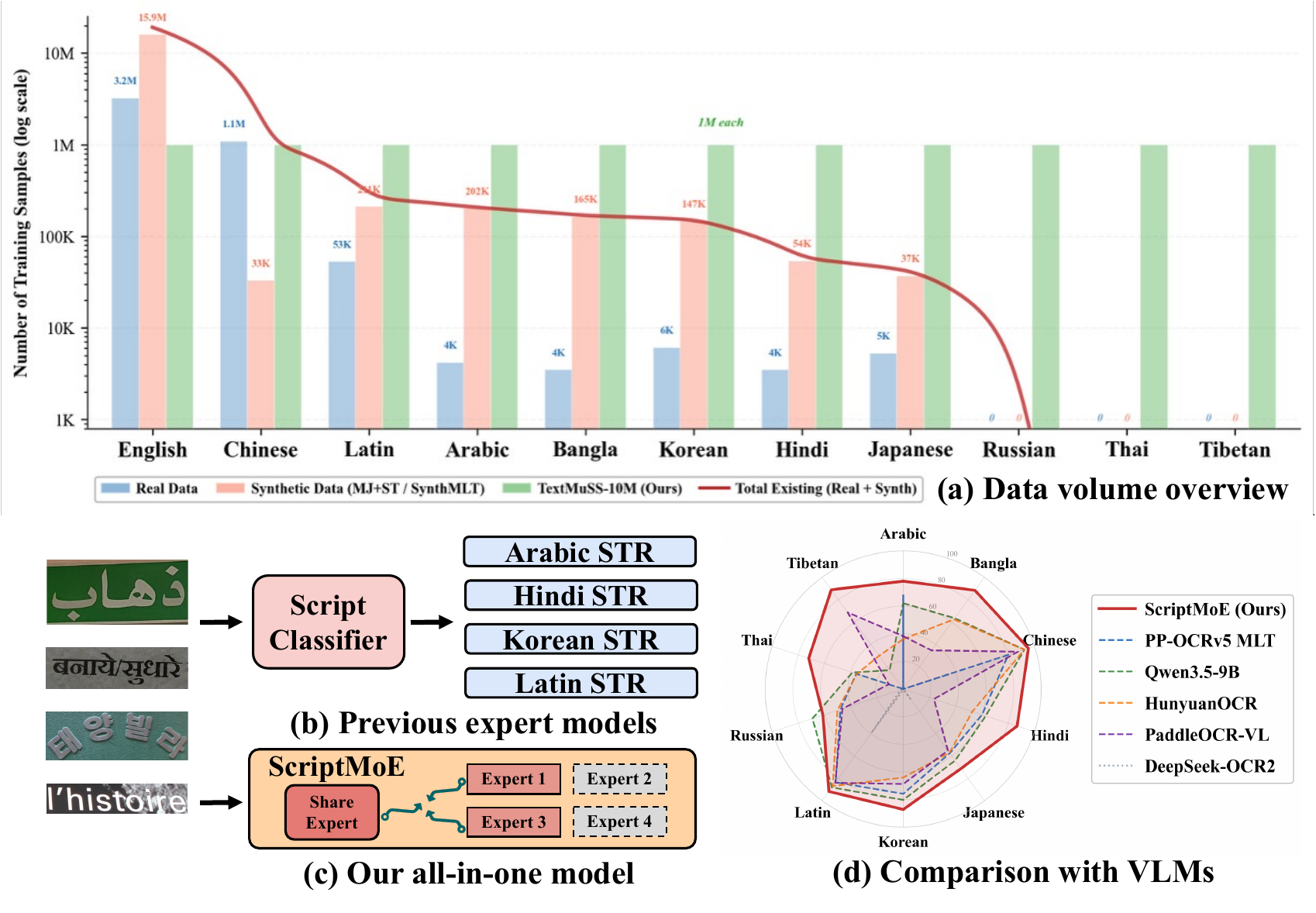} 
\caption{Current landscape of multilingual STR datasets (\textbf{a}). Existing datasets are heavily biased toward major languages, and some languages have no corresponding data available. Comparison with existing mainstream multilingual OCR systems in terms of methodology (\textbf{b, c}) and performance (\textbf{d}).}
\label{fig:teaser}
\end{figure}

\section{Introduction}
Scene text recognition (STR) is one of the most widely deployed OCR tasks, aiming to read text from cluttered natural images. Existing research has overwhelmingly centered on high-resource languages such as English and Chinese, where, driven by ever-stronger models and data, accuracy is now approaching saturation~\cite{li2023trocr, du2024svtrv2, ye2026wrong}. Once we step outside this comfort zone, however, the problem is far from settled. Real-world OCR systems must read text and symbols in many more scripts like Japanese, Korean, Arabic, Hindi, Cyrillic, and beyond. Today's multilingual OCR approaches fall broadly into two camps. Expert OCR systems~\cite{cui2025paddleocr} cascade a scene text detector with multilingual STR models. Typically they still deploy one recognizer per language and require an extra language-identification step to route each crop to the language-specific recognizer. This inflates training and deployment cost, complicates maintenance, and introduces error accumulation: once the language classifier is wrong, no downstream model can recover. VLM-based systems~\cite{cui2025paddleocrvl, team2025hunyuanocr, unirec, li2025dots, dong2026qianfan, team2026qwen3} unify many languages in a general or OCR-specialized visual language model, yet their massive parameter counts and inference cost make edge deployment difficult.  We therefore set out to build an all-in-one multilingual recognizer. It is a single model that is simpler than per-language experts, lighter than VLMs, and more accurate than both (see the per-script performance in Fig.~\ref{fig:teaser} (d)).

The world has hundreds of languages, but grouping them by script collapses this number dramatically. PP-OCRv5 MLT~\cite{cui2025paddleocr}, for instance, covers 106 languages with only about ten per-script models. This script-centric view greatly simplifies modeling a unified multilingual recognizer. We accordingly consolidate the major world languages into ten representative scripts: the most widespread Latin; the Latin-adjacent Cyrillic (e.g.\ Russian); the Han-derived but mutually distinct Chinese, Japanese and Korean; the right-to-left Arabic; and the Hindi family. To stress generality we further include the Southeast-Asian Thai, and the minority scripts Bangla and Tibetan. Modeling these ten scripts already covers 229 languages, and the full language-to-script mapping is given in the appendix.

Even after this consolidation, every script other than English and Chinese lacks large-scale real STR data for training~\cite{textssr}, as the dataset landscape in Fig.~\ref{fig:teaser} (a) makes clear. Following and expanding the synthetic engine UnionST~\cite{ye2026wrong}, we synthesize 1M samples for each script (\textbf{10M} in total) to form \textbf{TextMuSS-10M}, a \textbf{Mu}ltilingual \textbf{S}ynthetic \textbf{S}cene \textbf{Text} dataset. Although synthetic, it guarantees language coverage, balance and scene diversity, and is the most practical way to obtain large-scale supervision when real data is unavailable. Our overall training mixture is thus threefold: abundant existing real data for English (Union14M~\cite{jiang2023revisiting}) and Chinese (BCTR~\cite{bctr}), a small amount of real multilingual data (MLT2019~\cite{nayef2019icdar2019}), and the large-scale synthetic TextMuSS-10M covering all scripts.

For these ten scripts we assemble and re-collect real scene text images for evaluation: we reuse the MLT2019 test set (seven scripts) and additionally collect images for Tibetan, Russian and Thai. We refer to the resulting benchmark as \textbf{TextMuSS-Bench}, a \textbf{Mu}ltilingual ten-\textbf{S}cript \textbf{S}cene \textbf{Text} \textbf{Bench}mark. Simply training mainstream STR methods on our data already lets the jointly trained model surpass multilingual OCR expert systems and strong VLMs, a first confirmation that the all-in-one route is viable.

Yet joint training alone exposes a deeper problem. Scripts differ fundamentally in stroke topology (the stacked consonants of Hindi and Tibetan, the cursive ligatures of Arabic), reading direction (right-to-left Arabic), and script-internal vocabulary size (Han character sets are tens to hundreds of times larger than the Latin alphabet). Reusing a single dense recognizer originally designed for one language forces one parameter budget to be shared across these conflicting priors, with two consequences that pull against each other: (i) low-resource scripts (Thai, Tibetan) receive too little capacity, while (ii) common scripts are never fully specialized.

Our key observation is that a scene text instance always contains a single script (occasionally bilingual mix). Under this single-image-few-script prior, a Mixture-of-Experts (MoE) is an especially natural fit: the router nearly always faces an unambiguous decision. Incremental multilingual text recognition (IMLTR) presented by MRN~\cite{zheng2023mrn} explores routing-like ideas, but maintains a separate feature extractor per language and activates a language-specific expert with a distinct character classifier. So their parameters grow linearly with the number of languages. Inspired by how MoE ``activates expert parameters on demand'' in LLMs and VLMs~\cite{shazeer2017outrageously, switch,vmoe,deepseekmoe}, our multilingual recognizer shares one visual encoder followed by an MoE-based decoder whose experts are aligned with script clusters: an always-on shared expert absorbs cross-script knowledge, while the router activates the Top-2 script experts matching the current language. The resulting model, \textbf{ScriptMoE}, meets the all-in-one goal while keeping per-language performance mutually non-interfering.

Experimental results demonstrate the effectiveness of ScriptMoE across diverse scripts. On TextMuSS-Bench, it achieves state-of-the-art (SOTA) average accuracy, outperforming the jointly trained baseline by 1.31\%, with 2-3 points on Arabic, Thai and Tibetan. It even improves by about 18\% over PP-OCRv5 MLT and Qwen3.5-9B~\cite{team2026qwen3}. It also benefits end-to-end multilingual OCR: keeping the PP-OCRv5 detector and replacing only the recognizer with ScriptMoE raises PP-OCRv5 MLT's F1 score on the CC-OCR~\cite{ccocr} multilingual task from 65.71\% to 80.89\%, slightly outperforming the best VLM (80.73\%) despite the latter having a hundred times more parameters. This is a win on both efficiency and accuracy.

In summary, our main contributions are:

\begin{itemize}
\item We build TextMuSS-10M, a large-scale and balanced multilingual synthetic scene text dataset spanning 10 scripts and 229 languages, and assemble TextMuSS-Bench, a real scene  text benchmark covering these scripts.
\item We propose ScriptMoE, a script-aware MoE-based (Top-2 routed experts plus an always-on shared expert for cross-script transfer) recognizer with image-level routing and lightweight script-aware supervision.
\item We systematically evaluate all-in-one multilingual STR under a unified training protocol. ScriptMoE achieves the best average accuracy on TextMuSS-Bench. With the PP-OCRv5 detector, it performs on par with VLMs on end-to-end multilingual OCR while remaining lightweight.
\end{itemize}

\section{Related Work}

\paragraph{Scene Text Recognition.}
STR models can be split, by decoding strategy, into autoregressive (AR) and non-autoregressive (NAR) families. AR methods~\cite{Sheng2019nrtr, parseq, jiang2023revisiting, ccd, ccdplus, ye2026advancing} introduce explicit language modeling and lead in accuracy, but decode token-by-token and are comparatively slow.  NAR methods include classic CTC decoding~\cite{CTC,shi2017crnn,duijcai2022svtr,du2024svtrv2} and purpose-built parallel decoding~\cite{abinet,cppd,yang2025ipad}: few-step decoding is fast but lacks expressive language priors. Crucially, both families are designed for the monolingual English/Chinese setting and are rarely adapted to other scripts. In the multilingual regime their weaknesses surface: NAR models, being vision-centric and prior-poor, degrade further on morphologically rich scripts such as Hindi and Tibetan, and are ill-suited to joint multi-script recognition. Directly reusing an AR model, in contrast, lets all scripts contend for one dense set of decoder parameters and imbalances high- and low-resource languages. We therefore keep an AR backbone but reallocate its decoder capacity to be script-aware, activating only the expert parameters for the current script.

\paragraph{Multilingual OCR.}
The community first approaches multilingual scene text through competitions~\cite{nayef2017icdar2017,nayef2019icdar2019}, which laid their data and evaluation foundations. E2E-MLT~\cite{e2e-mlt} builds the first systematic end-to-end multilingual OCR pipeline and contributes the SynthMLT. On how to handle multiple scripts, Multiplexed TextSpotter~\cite{huang2021multiplexed} performs word-level script identification and routes each word to a script-specific recognition head. SARN~\cite{ke2024sarn} injects script information into the recognizer to make character features more discriminative. For the resource-constrained incremental setting, MRN~\cite{zheng2023mrn} formulates IMLTR~\cite{hammer, crossknow, risen} with a language-domain router. On the data side, CLI-STR~\cite{baek2024cross} finds that data scale, not linguistic similarity, is the decisive factor. It motivates our large-scale synthetic data for low-resource scripts. Among deployed systems, general multilingual OCR~\cite{cui2025paddleocr} maintains a separate model per script and must be told the target language in advance. Recent generalist~\cite{team2026qwen3, wang2025internvl3} and OCR-specific VLMs~\cite{cui2025paddleocrvl, team2025hunyuanocr, li2025dots, dong2026qianfan} claim to transcribe many scripts. But, as our experiments show, their performance remains markedly behind specialized recognizers on one or more scripts. This again argues for a lighter, unified, high-accuracy recognizer.

\section{Method}
\label{sec:method}

\subsection{Overview}
\begin{figure}[t]
\centering
\includegraphics[width=0.8\textwidth]{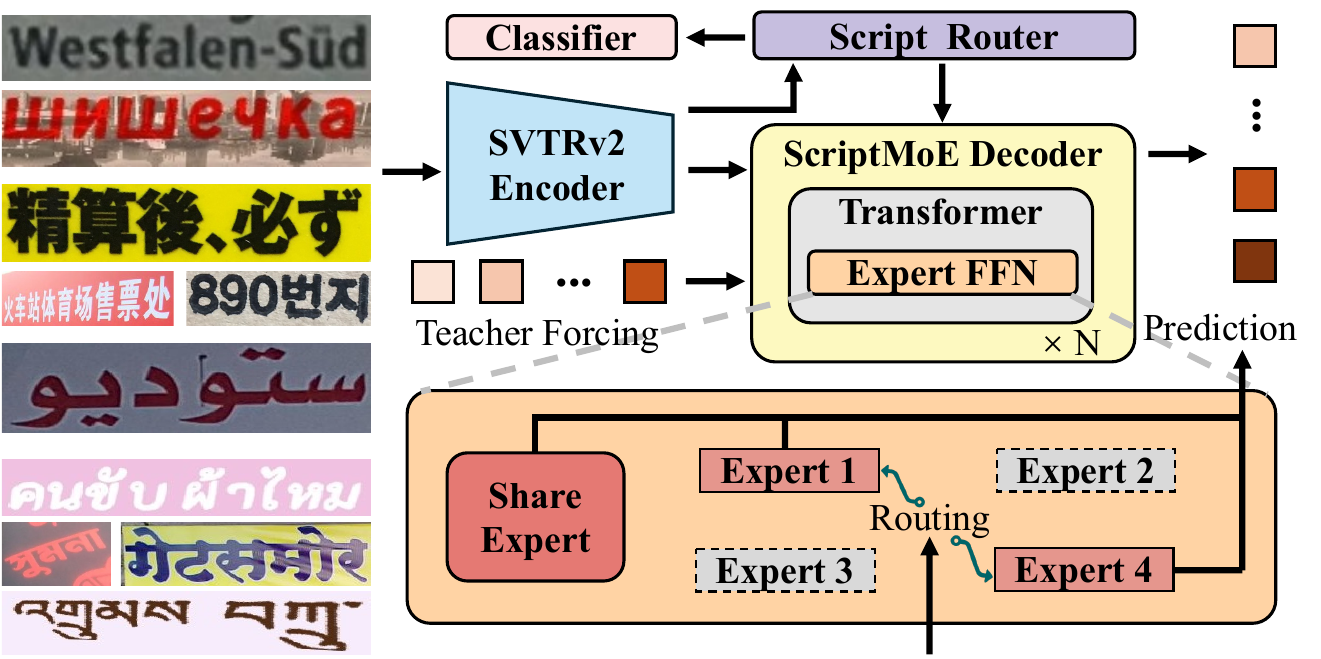} 
\caption{Overview of ScriptMoE. Multilingual scene text (real and synthetic mixed) is encoded by a visual encoder and fed into a decoder that replaces the FFN with an image-level script-aware MoE block. The script  router selects the top-2 script experts per image from pooled visual features, and all output tokens share this route. Experts have explicit script-family roles, while the shared expert is always active. After N=2 such modified Transformer blocks, it passes through a classification head to produce the final prediction. The training process follows the standard teacher-forcing paradigm.
}
\label{fig:arch}
\end{figure}

We target a single network that transcribes multilingual text with high accuracy. So we identify two key obstacles to building such a recognizer. \textbf{(i)~Data.} Existing STR supervision is overwhelmingly in Chinese and English, leaving most long-tail scripts impossible to learn from real data. We therefore construct TextMuSS-10M (see examples in Fig.~\ref{fig:data}), a balanced multilingual synthetic dataset that provides every script with a comparable amount of training signal. \textbf{(ii)~Capacity.} Even with data in hand, a single dense decoder must amortize one parameter budget across scripts. This both blurs script-specific features and starves low-resource scripts. Building on the single-image-few-script prior, we replace its dense FFN with ScriptMoE, which makes one routing decision per image rather than per token. Its router and auxiliary script classifier are separate heads sharing only the pooled visual representation, so the classifier supervises script-aware specialization without directly selecting experts. Each routed expert has an explicit script-family role, while an always-on shared expert preserves cross-script transfer. Fig.~\ref{fig:arch} sketches the full pipeline.

\subsection{Balanced Multilingual Data through Synthesis}
SynthMLT~\cite{e2e-mlt} is the only multilingual synthetic STR dataset currently available. We empirically find that its scale and quality are too limited to support high-accuracy model training. In addition, the languages it covers are the ten mentioned in MLT2019~\cite{nayef2019icdar2019}. This cannot meet our need to support hundreds of languages. Therefore, we need to build our own large, high-quality synthetic dataset. It will serve as the foundation for subsequent script-balanced, high-accuracy models.

\begin{figure}[t]
\centering
\includegraphics[width=0.9\textwidth]{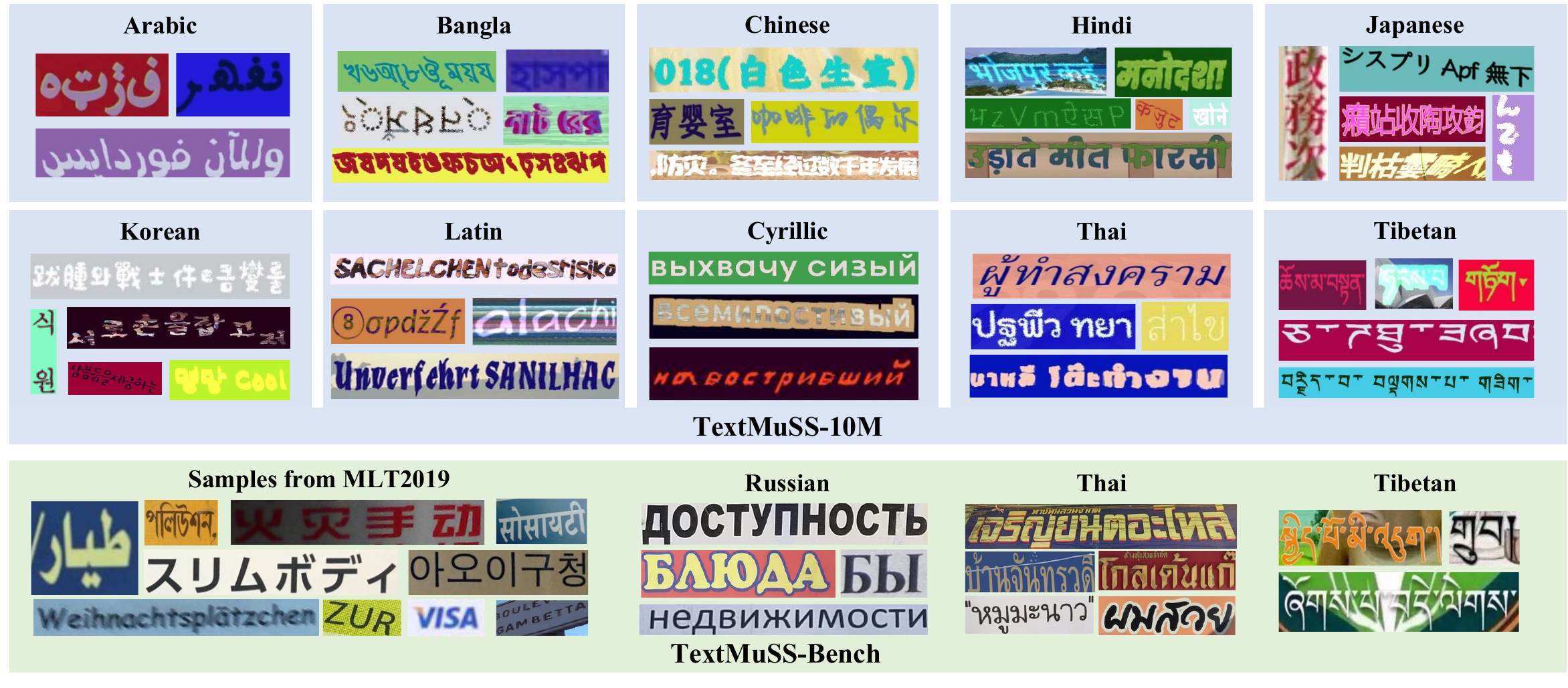}
\caption{\textbf{Top:} Data samples from TextMuSS-10M, organized by script. \textbf{Bottom:} Representative samples from TextMuSS-Bench, which extends the original MLT2019 with newly collected real-world samples in three additional languages for evaluation.}
\label{fig:data}
\end{figure}

We follow the current strong STR synthesis engine, UnionST~\cite{ye2026wrong}. We adapt it to our task in the following ways. (1) Character vocabulary. We first collect the character set to be supported for each of the ten target scripts. We refer to PP-OCRv5 MLT~\cite{cui2025paddleocr} and the standard definition of each script. This character set is then used for filtering and sampling expansion in subsequent data. (2) Corpus collection. The scene text corpus is word-level. So we collect 100K to 1M words for each script. For scripts like Latin that cover dozens of languages, we collect more words and try to balance them across languages. To cover longer text, we concatenate the words above with spaces. This simulates phrase- and line-level corpora. To further balance the length and character distributions and to cover rare characters, we add meaningless text generated by random permutations of the character table. To simulate sentence-level real semantics, we collect newspaper corpora for our target languages from News Crawl~\cite{barrault2019findings}. We then extract phrases and sentences of various lengths from them. The result is a multilingual corpus with comprehensive coverage and balanced distribution. (3) Detail adjustments for language features. Chinese, Japanese, and Korean need a larger proportion of vertical-text synthesis (20\%, while the rest is 5\%). The synthesis of texts such as Arabic needs to be rendered from right to left (saved in logical order).

The specific synthesis flow is as follows. (1) We select 8k text-free scene images or pure-color images as the background layer. (2) We select a text from the corresponding language corpus. (3) We use a rendering tool such as Pillow to render the text according to preset layout templates. The templates include horizontal, vertical, multi-directional rotation, and curved. This produces a text layer. (4) We add effects (shadow, distortion, perspective) to the text layer. We then overlay it on the background layer. The above process produces the synthesis effect shown in the top of Fig.~\ref{fig:data}. It constitutes 1M synthetic data per script, TextMuSS-10M.

\subsection{Script-aware Mixture-of-Experts}
\label{sec:moe}

With data ready, the next obstacle is architectural. Our enabling insight is the single-image-few-script prior. The recognizer therefore does not need a dense decoder that is simultaneously fluent in every script. It needs the ability to switch to the right specialist. Sparsely-activated experts express exactly this.

Specifically, ScriptMoE starts with a strong and hierarchical visual encoder (SVTRv2~\cite{du2024svtrv2} here). It maps an input image $x \in \mathbb{R}^{H \times W \times 3}$ to a sequence of visual tokens $\mathbf{F} \in \mathbb{R}^{L \times d}$. These tokens are then fed into a script-aware Transformer decoder, in which the feed-forward network (FFN) is replaced by an MoE-based FFN. The remaining components stay identical to a vanilla Transformer decoder that produces the output autoregressively step by step. Let $\mathbf{h}_t \in \mathbb{R}^{d}$ be the decoder hidden state at step $t$, taken after the attention sub-layer and before the FFN. With $n$ routed experts and one always-on shared expert, the layer computes: \begin{equation} \mathrm{MoE}(\mathbf{h}_t) = \alpha_t\, \mathrm{FFN}_{\text{share}}(\mathbf{h}_t) + (1-\alpha_t)\sum_{i=1}^{n} g_i\, \mathrm{FFN}_i(\mathbf{h}_t), \label{eq:moe} \end{equation} \begin{equation} g_i = \frac{p_i\cdot \mathbb{I}\{i\in \mathrm{TopK}(\mathbf{p})\}}{\sum_{j} p_j\cdot \mathbb{I}\{j\in \mathrm{TopK}(\mathbf{p})\}}, \label{eq:router} \end{equation} \begin{equation} \mathbf{p} = \mathrm{softmax}\!\bigl(W_g(\bar{\mathbf{F}}\odot(1 + \sigma\cdot \boldsymbol{\epsilon}))\bigr), \label{eq:softmax} \end{equation} where $\bar{\mathbf{F}} \in \mathbb{R}^{d}$ is the image-level router input, obtained by mean-pooling the visual tokens $\mathbf{F}$. $\boldsymbol{\epsilon}\sim\mathcal{N}(0, I)$ is a multiplicative router jitter applied only during training with $\sigma$ as the jitter coefficient, $W_g$ is the router projection, and $\mathrm{TopK}$ keeps the highest-scoring experts (Top-2 here), whose gates $g_i$ are renormalized to sum to one. Instead of a fixed weight, a learnable per-token gate $\alpha_t=\mathrm{sigmoid}(\mathbf{w}_s^{\top}\mathbf{h}_t)$ balances the shared and routed branches. The resulting $\mathrm{MoE}(\mathbf{h}_t)$ directly replaces the FFN output of the vanilla Transformer.

\paragraph{Script Grouping.}
Based on the character morphology similarity of the ten scripts, we further divide them into four groups to save expert capacity and achieve optimal performance. The first group is the alphabet group, including Latin + Cyrillic. The second group is CJK, including Chinese, Japanese, and Korean. The third group is the Arabic family. The fourth group is Others, including Hindi, Bangla, Tibetan, and Thai. We also reserve sufficient room for unsupported languages. If a script is visually similar to letters, it can be assigned to the first group. If it is in Han-character form, it can be assigned to the second group. If it follows an RTL reading order, it falls into the third group. If none of the above applies, it can be assigned to the fourth group. The subsequent routing mechanism is also built upon this grouping strategy. Crucially, the router input $\bar{\mathbf{F}}$ is formed once per image rather than once per token, so all $T$ output tokens of an image are processed by the same routed experts. This is the natural encoding of the single-image-few-script prior and brings three structural advantages: it (a) cuts routing cost, (b) avoids the token-level ``ping-ponging'' that destabilizes training, and (c) makes each expert directly readable as a script specialist.

\paragraph{Shared Expert.}
The shared expert $\mathrm{FFN}_{\text{share}}$ is always activated, regardless of the routing decision, and is the one path through which gradients from every image flow. Its role is to absorb the script-invariant part of the problem: symbols that recur across scripts (digits, punctuation), the geometry of curved and perspective-distorted text, and vocabulary shared within close families such as Chinese and Japanese. It is what preserves cross-script transfer, so that specializing the routed experts does not fragment the common sense shared across all scripts.

\subsection{Script-aware Supervision}
\paragraph{Script-classification Signal.}
Left unsupervised, the router is free to discover any grouping of images, which need not align with scripts. We give it a gentle nudge with a four-way classification head $h_{\text{scls}}:\mathbb{R}^{d}\to \mathbb{R}^{4}$ attached to the same router input $\bar{\mathbf{F}}$, and trained with the cross-entropy between its softmax output and the script-group label $y_x$, which is derived automatically by mapping the Unicode ranges of the ground-truth transcription characters to one of the four script groups,
\begin{equation}
\mathcal{L}_{\text{scls}} = -\frac{1}{|\mathcal{B}|}\sum_{x\in\mathcal{B}}
\log \mathrm{softmax}\!\big(h_{\text{scls}}(\bar{\mathbf{F}}_x)\big)_{y_x},
\label{eq:scls}
\end{equation}
where $\mathcal{B}$ is the mini-batch. Importantly, $h_{\text{scls}}$ shares the router input but not the router weights: it supplies a script-aware learning signal while still letting the router carve out useful within-script sub-populations.

\paragraph{Training Objective.}
ScriptMoE is trained end-to-end by minimizing
\begin{equation}
\mathcal{L} = \mathcal{L}_{\text{ar}} + \lambda_{\text{scls}}
\mathcal{L}_{\text{scls}},
\end{equation}
where $\mathcal{L}_{\text{ar}}$ is the standard autoregressive cross-entropy and dominates the objective. The single auxiliary term $\mathcal{L}_{\text{scls}}$ only shapes how the experts are used and is therefore weighted far below it. We set $\lambda_{\text{scls}}{=}0.1$, enough to orient the experts towards scripts but not so strong as to override the router's freedom.

\begin{table}[t]
\centering
\caption{Per-script size of TextMuSS-Bench. Latin includes English, French,
German, Italian, etc.}
\label{tab:benchA_size}
\setlength{\tabcolsep}{1.5mm}
\begin{tabular}{l*{10}{c}|c}
\hline
Script & Arabic & Bangla & Chinese & Hindi & Japanese & Korean & Latin & Russian & Thai & Tibetan & \textbf{Total} \\
\hline
\# img & 470 & 393 & 325 & 393 & 594 & 679 & 5,885 & 1,054 & 750 & 356 & \textbf{10,899} \\
\hline
\end{tabular}
\end{table}

\begin{table*}
\small
\caption{Performance comparison on TextMuSS-Bench. Results are reported as a / b, where a is word accuracy (\%) and b is 1 - normalized edit distance (. denotes 0.). Avg is the arithmetic average across ten scripts. For PP-OCRv5 MLT~\cite{cui2025paddleocr}, it is computed over the supported ones. For each metric, best in \textbf{bold}, second \underline{underlined}. ``/'' indicates the model does not support this language. }
\label{tab:main_mlt}
\resizebox{\textwidth}{!}{%
\begin{tabular}{l|cccccccccc|c}
\hline
Method & Arabic & Bangla & Chinese & Hindi & Japanese & Korean & Latin & Russian & Thai & Tibetan & Avg \\
\hline
\multicolumn{12}{c}{\textit{General / OCR-specialized VLMs (zero-shot)}}\\
InternVL3.5-8B~\shortcite{wang2025internvl3} & {\scriptsize 0.00~/~.09} & {\scriptsize 0.00~/~.03} & {\scriptsize 72.31~/~.82} & {\scriptsize 0.25~/~.09} & {\scriptsize 32.32~/~.55} & {\scriptsize 38.88~/~.50} & {\scriptsize 74.34~/~.83} & {\scriptsize 30.93~/~.48} & {\scriptsize 0.40~/~.09} & {\scriptsize 0.00~/~.00} & {\scriptsize 24.94~/~.35} \\
Qwen3.5-9B~\shortcite{team2026qwen3} & {\scriptsize 62.13~/~.78} & {\scriptsize 63.36~/~.78} & {\scriptsize 92.31~/~.94} & {\scriptsize 62.85~/~.82} & {\scriptsize 63.97~/~.81} & {\scriptsize 80.27~/~.87} & {\scriptsize 88.04~/~.84} & {\scriptsize \textbf{69.17}~/~\textbf{.74}} & {\scriptsize 38.76~/~.65} & {\scriptsize 16.85~/~.74} & {\scriptsize 63.77~/~.80} \\
GOT-OCR 2.0~\shortcite{wei2024general} & {\scriptsize 0.00~/~.00} & {\scriptsize 0.00~/~.02} & {\scriptsize 76.92~/~.86} & {\scriptsize 0.25~/~.02} & {\scriptsize 21.72~/~.44} & {\scriptsize 0.59~/~.04} & {\scriptsize 84.71~/~.92} & {\scriptsize 0.47~/~.09} & {\scriptsize 0.00~/~.04} & {\scriptsize 0.00~/~.00} & {\scriptsize 18.47~/~.24} \\
HunyuanOCR~\shortcite{team2025hunyuanocr} & {\scriptsize 36.17~/~.44} & {\scriptsize 62.34~/~.67} & {\scriptsize 92.31~/~.93} & {\scriptsize 52.67~/~.62} & {\scriptsize 57.07~/~.72} & {\scriptsize 64.06~/~.68} & {\scriptsize 86.85~/~.90} & {\scriptsize 50.19~/~.61} & {\scriptsize 35.33~/~.53} & {\scriptsize 29.78~/~.42} & {\scriptsize 56.68~/~.65} \\
PaddleOCR-VL~\shortcite{cui2025paddleocrvl} & {\scriptsize 38.30~/~.63} & {\scriptsize 34.61~/~.55} & {\scriptsize 87.38~/~.92} & {\scriptsize 23.41~/~.49} & {\scriptsize 55.05~/~.76} & {\scriptsize 68.78~/~.78} & {\scriptsize 83.89~/~.90} & {\scriptsize 46.30~/~.62} & {\scriptsize 11.07~/~.34} & {\scriptsize 68.54~/~.89} & {\scriptsize 51.73~/~.69} \\
DeepSeek-OCR2~\shortcite{wei2026deepseek} & {\scriptsize 0.00~/~.01} & {\scriptsize 0.00~/~.01} & {\scriptsize 35.38~/~.44} & {\scriptsize 0.25~/~.01} & {\scriptsize 8.59~/~.21} & {\scriptsize 0.59~/~.02} & {\scriptsize 39.29~/~.50} & {\scriptsize 1.99~/~.08} & {\scriptsize 0.00~/~.02} & {\scriptsize 0.00~/~.01} & {\scriptsize 8.61~/~.13} \\
GLM-OCR~\shortcite{duan2026glm} & {\scriptsize 1.49~/~.22} & {\scriptsize 3.31~/~.29} & {\scriptsize 90.46~/~.94} & {\scriptsize 9.14~/~.41} & {\scriptsize 56.40~/~.77} & {\scriptsize 40.06~/~.62} & {\scriptsize 89.62~/~.94} & {\scriptsize 50.95~/~.69} & {\scriptsize 0.13~/~.13} & {\scriptsize 0.00~/~.01} & {\scriptsize 34.16~/~.50} \\
\multicolumn{12}{c}{\textit{Expert OCR Systems (zero-shot)}}\\
PP-OCRv5 MLT~\shortcite{cui2025paddleocr} & {\scriptsize 68.30~/~.86} & / & {\scriptsize 80.00~/~.86} & {\scriptsize 60.05~/~.84} & {\scriptsize 57.41~/~.76} & {\scriptsize 75.85~/~.84} & {\scriptsize 83.43~/~.91} & {\scriptsize 47.82~/~.67} & {\scriptsize 35.87~/~.69} & / & {\scriptsize 63.59~/~.80} \\
PP-OCRv6~\shortcite{zhang2026pp} & / & / & {\scriptsize 84.62~/~.89} & / & {\scriptsize 58.75~/~.75} & / & {\scriptsize 88.05~/~.93} & / & / & / & {\scriptsize /} \\
\hline
\multicolumn{12}{c}{\textit{NAR-based (CTC / PD) STR}}\\
CRNN~\shortcite{shi2017crnn} & {\scriptsize 42.77~/~.72} & {\scriptsize 55.47~/~.82} & {\scriptsize 64.00~/~.80} & {\scriptsize 63.87~/~.86} & {\scriptsize 44.28~/~.66} & {\scriptsize 70.10~/~.82} & {\scriptsize 82.38~/~.90} & {\scriptsize 30.93~/~.44} & {\scriptsize 53.13~/~.72} & {\scriptsize 48.60~/~.85} & {\scriptsize 55.55~/~.76} \\
ABINet~\shortcite{abinet} & {\scriptsize 61.06~/~.84} & {\scriptsize 69.47~/~.90} & {\scriptsize 82.15~/~.94} & {\scriptsize 79.90~/~.94} & {\scriptsize 58.08~/~.83} & {\scriptsize 81.74~/~.90} & {\scriptsize 88.84~/~\underline{.95}} & {\scriptsize 50.76~/~.46} & {\scriptsize 53.20~/~.82} & {\scriptsize 73.88~/~.92} & {\scriptsize 69.91~/~.85} \\
SVTR~\shortcite{duijcai2022svtr} & {\scriptsize 64.68~/~.83} & {\scriptsize 75.83~/~\underline{.92}} & {\scriptsize 83.38~/~.94} & {\scriptsize 81.42~/~.93} & {\scriptsize 60.44~/~.81} & {\scriptsize 83.95~/~.91} & {\scriptsize 90.06~/~\underline{.95}} & {\scriptsize 52.75~/~.46} & {\scriptsize 58.80~/~.82} & {\scriptsize 78.65~/~.93} & {\scriptsize 73.00~/~.85} \\
LPV~\shortcite{lpv} & {\scriptsize 66.17~/~.88} & {\scriptsize 74.30~/~\textbf{.94}} & {\scriptsize 84.00~/~.94} & {\scriptsize 82.44~/~.95} & {\scriptsize 60.27~/~.85} & {\scriptsize 82.92~/~\underline{.92}} & {\scriptsize 90.03~/~\underline{.95}} & {\scriptsize 49.81~/~.36} & {\scriptsize 56.80~/~.80} & {\scriptsize 74.72~/~.91} & {\scriptsize 72.15~/~.85} \\
BUSNet~\shortcite{busnet} & {\scriptsize 61.49~/~.78} & {\scriptsize 70.48~/~.82} & {\scriptsize 85.54~/~.90} & {\scriptsize 77.10~/~.86} & {\scriptsize 61.62~/~.75} & {\scriptsize 83.51~/~.89} & {\scriptsize 90.18~/~.93} & {\scriptsize 49.43~/~.71} & {\scriptsize 53.87~/~.74} & {\scriptsize 74.44~/~.85} & {\scriptsize 70.77~/~.82} \\
CPPD~\shortcite{cppd} & {\scriptsize 67.02~/~.81} & {\scriptsize 78.12~/~.87} & {\scriptsize 78.77~/~.85} & {\scriptsize 79.39~/~.87} & {\scriptsize 57.24~/~.73} & {\scriptsize 81.74~/~.88} & {\scriptsize 88.36~/~.92} & {\scriptsize 40.13~/~.67} & {\scriptsize 52.53~/~.72} & {\scriptsize 73.60~/~.85} & {\scriptsize 69.69~/~.82} \\
SVTRv2~\shortcite{du2024svtrv2} & {\scriptsize 70.85~/~.86} & {\scriptsize 83.46~/~\underline{.92}} & {\scriptsize 93.23~/~\underline{.97}} & {\scriptsize 84.73~/~.94} & {\scriptsize 66.67~/~.85} & {\scriptsize 85.86~/~\underline{.92}} & {\scriptsize 91.52~/~\underline{.95}} & {\scriptsize 47.91~/~.41} & {\scriptsize 66.27~/~.83} & {\scriptsize 85.39~/~\underline{.95}} & {\scriptsize 77.59~/~.86} \\
MDiff4STR~\shortcite{du2026mdiff4str} & {\scriptsize 72.13~/~.84} & {\scriptsize 83.46~/~.90} & {\scriptsize 88.31~/~.92} & {\scriptsize 83.21~/~.90} & {\scriptsize 66.50~/~.79} & {\scriptsize 85.13~/~.90} & {\scriptsize 91.03~/~.94} & {\scriptsize 42.31~/~.68} & {\scriptsize 57.33~/~.75} & {\scriptsize 75.84~/~.86} & {\scriptsize 74.53~/~.85} \\
\multicolumn{12}{c}{\textit{AR-based STR}}\\
NRTR~\shortcite{Sheng2019nrtr} & {\scriptsize 67.66~/~.84} & {\scriptsize 79.64~/~.89} & {\scriptsize 81.23~/~.91} & {\scriptsize 83.21~/~.92} & {\scriptsize 60.44~/~.80} & {\scriptsize 81.15~/~.89} & {\scriptsize 89.97~/~.94} & {\scriptsize 58.53~/~.45} & {\scriptsize 59.39~/~.82} & {\scriptsize 71.07~/~.91} & {\scriptsize 73.23~/~.84} \\
SEED~\shortcite{qiao2020seed} & {\scriptsize 66.60~/~.80} & {\scriptsize 80.92~/~.89} & {\scriptsize 84.62~/~.90} & {\scriptsize 83.46~/~.90} & {\scriptsize 62.12~/~.75} & {\scriptsize 82.77~/~.88} & {\scriptsize 90.64~/~.94} & {\scriptsize 51.80~/~\underline{.72}} & {\scriptsize 56.53~/~.76} & {\scriptsize 76.40~/~.87} & {\scriptsize 73.59~/~.84} \\
PARSeq~\shortcite{parseq} & {\scriptsize 67.87~/~.87} & {\scriptsize 78.12~/~\underline{.92}} & {\scriptsize 87.08~/~\underline{.97}} & {\scriptsize 83.21~/~.95} & {\scriptsize 65.15~/~.85} & {\scriptsize 84.98~/~\underline{.92}} & {\scriptsize 90.81~/~\textbf{.96}} & {\scriptsize 61.01~/~.47} & {\scriptsize 62.67~/~.82} & {\scriptsize 84.55~/~\underline{.95}} & {\scriptsize 76.55~/~.87} \\
MAERec~\shortcite{jiang2023revisiting} & {\scriptsize \underline{76.17}~/~.85} & {\scriptsize 80.15~/~.87} & {\scriptsize 92.92~/~.95} & {\scriptsize 86.01~/~.90} & {\scriptsize 68.35~/~.78} & {\scriptsize \underline{86.16}~/~.90} & {\scriptsize \underline{92.13}~/~.94} & {\scriptsize 52.66~/~.71} & {\scriptsize 63.73~/~.77} & {\scriptsize \underline{86.80}~/~.91} & {\scriptsize 78.51~/~.86} \\
CDistNet~\shortcite{zheng2024cdistnet} & {\scriptsize 71.28~/~.87} & {\scriptsize 82.95~/~\underline{.92}} & {\scriptsize 89.23~/~.96} & {\scriptsize 85.50~/~.95} & {\scriptsize 67.51~/~\underline{.86}} & {\scriptsize 85.42~/~.91} & {\scriptsize 91.18~/~\underline{.95}} & {\scriptsize 59.49~/~.45} & {\scriptsize 64.13~/~.83} & {\scriptsize 82.87~/~.94} & {\scriptsize 77.96~/~.86} \\
SMTR~\shortcite{smtr} & {\scriptsize 70.43~/~.83} & {\scriptsize 77.86~/~.91} & {\scriptsize 88.92~/~.96} & {\scriptsize 85.24~/~.94} & {\scriptsize 65.32~/~.84} & {\scriptsize 85.27~/~.91} & {\scriptsize 91.06~/~\underline{.95}} & {\scriptsize 51.99~/~.44} & {\scriptsize 62.93~/~.84} & {\scriptsize 82.87~/~.93} & {\scriptsize 76.19~/~.86} \\
SVTRv2-AR~\shortcite{ye2026wrong} & {\scriptsize 75.11~/~\underline{.91}} & {\scriptsize \underline{87.79}~/~\textbf{.94}} & {\scriptsize \underline{94.15}~/~\underline{.97}} & {\scriptsize \textbf{87.28}~/~\underline{.96}} & {\scriptsize \underline{69.36}~/~\textbf{.87}} & {\scriptsize 85.86~/~\textbf{.93}} & {\scriptsize \textbf{92.27}~/~\underline{.95}} & {\scriptsize 59.30~/~.45} & {\scriptsize \underline{69.60}~/~\underline{.85}} & {\scriptsize \underline{86.80}~/~\underline{.95}} & {\scriptsize \underline{80.75}~/~\underline{.88}} \\
\textbf{ScriptMoE (ours)} & {\scriptsize \textbf{78.09}~/~\textbf{.92}} & {\scriptsize \textbf{88.30}~/~\textbf{.94}} & {\scriptsize \textbf{95.38}~/~\textbf{.98}} & {\scriptsize \underline{86.77}~/~\textbf{.97}} & {\scriptsize \textbf{71.21}~/~\textbf{.87}} & {\scriptsize \textbf{87.19}~/~\textbf{.93}} & {\scriptsize 91.67~/~\textbf{.96}} & {\scriptsize \underline{61.20}~/~~\underline{.72}} & {\scriptsize \textbf{72.00}~/~\textbf{.86}} & {\scriptsize \textbf{88.76}~/~\textbf{.96}} & {\scriptsize \textbf{82.06}~/~\textbf{.91}} \\
\hline
\end{tabular}
}
\end{table*}

\section{Experiments}

\subsection{Evaluation and Implementation Details}
We assess ScriptMoE along two complementary axes. The first is STR evaluation on TextMuSS-Bench, which we build by extending the MLT2019~\cite{nayef2019icdar2019} test split with Russian, Thai and Tibetan. This covers all ten major scripts on which ScriptMoE is trained, for a total of 10,899 images whose per-script sizes are listed in Tab.~\ref{tab:benchA_size}. The second is end-to-end multilingual OCR task on CC-OCR~\cite{ccocr}. We keep the PP-OCRv5 detector fixed and replace only the recognizer, so that any difference is attributable to recognition alone. To keep the main text focused, we further evaluate on two benchmarks in the appendix: Chinese BCTR~\cite{bctr} and English Union14M-Benchmark~\cite{jiang2023revisiting}. For a fair comparison, all STR baselines are trained on the same data for the same number of epochs as ScriptMoE and re-evaluated on TextMuSS-Bench by us. The generalist OCR and VLM systems are evaluated zero-shot through their official models. Other details are reported in the appendix.

\begin{table*}[t]
\centering
\small
\caption{End-to-end OCR on the CC-OCR multilingual task (F1 score, \%).  ``-'' means that a system whose per-language scores are not reported by QianfanOCR~\cite{dong2026qianfan}.}
\label{tab:ccocr}
\resizebox{\textwidth}{!}{%
\begin{tabular}{l|cccccccccc|c}
\toprule
Method & Korean & Japanese & Vietnamese & French & German & Italian & Spanish & Portuguese & Russian & Arabic & Total \\
\midrule
\multicolumn{12}{c}{\textit{General VLMs}}\\
Claude-3.5-Sonnet & 58.72 & 56.90 & /    & 76.76 & 70.66 & 68.38 & 73.42 & 76.07 & 51.01 & 65.29 & 65.68 \\
GPT-4o    & 74.20 & 66.96 & 70.11 & 81.17 & 73.60 & 69.01 & 78.95 & 80.90 & 67.22 & 72.31 & 73.44 \\
Gemini-1.5-Pro    & 80.01 & 73.52 & \underline{78.49} & 83.33 & 78.11 & 75.77 & 81.28 & 83.46 & 69.99 & 85.70 & 78.97 \\
Qwen2.5-VL-72B~\shortcite{bai2025qwen25vltechnicalreport}    & 85.36 & 76.27 & 78.16 & \underline{83.56} & 79.27 & \underline{77.81} & \underline{82.14} & \underline{83.65} & 71.09 & 79.44 & 79.68 \\
InternVL3.5-8B~\shortcite{wang2025internvl3}    & 54.96 & 65.85 & 55.69 & 79.53 & 42.16 & 64.94 & 80.03 & 76.34 & 42.13 & 38.33 & 60.00 \\
Qwen3.5-9B~\shortcite{team2026qwen3}       & 79.03 & 75.17 & \textbf{81.43} & 82.83 & 76.82 & \textbf{78.77} & \textbf{85.08} & \textbf{87.24} & \underline{78.97} & 82.01 & \underline{80.73} \\
GPT-5.6-Terra          & 74.93 & 72.60 & 66.98 & 78.63 & 73.39 & 71.74 & 76.96 & 78.77 & 63.49 & 59.79 & 71.73 \\
Gemini-3.5-Flash       & \underline{89.15} & \underline{85.89} & 73.17 & \textbf{84.84} & \textbf{81.43} & 74.07 & 81.35 & 80.07 & 67.21 & \underline{88.05} & 80.52 \\
\midrule
\multicolumn{12}{c}{\textit{OCR-specialized VLMs}}\\
KOSMOS2.5~\shortcite{lv2023kosmos} & 26.95 & 26.94 & 27.24 & 48.93 & 47.04 & 44.59 & 52.92 & 54.56 & 11.13 & 22.04 & 36.23 \\
GOT-OCR 2.0~\shortcite{wei2024general}  & 27.53 & 43.66 &  7.92 & 33.74 & 28.73 & 28.53 & 31.63 & 25.20 &  3.85 & 18.72 & 24.95 \\
DeepSeek-OCR~\shortcite{wei2025deepseek}      & - & - & - & - & - & - & - & - & - & - & 32.50 \\
Dots.ocr~\shortcite{li2025dots}  & - & - & - & - & - & - & - & - & - & - & 47.20 \\
Surya OCR 2 ~\shortcite{paruchuri2025surya} & 70.35 & 71.45 & 63.49 & 63.79 & 57.62 & 51.90 & 	69.26 & 69.31 & 55.88 & 76.30 & 64.94 \\
MinerU2.5~\shortcite{niu2026mineru2} & - & - & - & - & - & - & - & - & - & - & 43.20 \\
PaddleOCR-VL~\shortcite{cui2025paddleocrvl}      & - & - & - & - & - & - & - & - & - & - & 45.50 \\
Qianfan-OCR~\shortcite{dong2026qianfan}       & - & - & - & - & - & - & - & - & - & - & 76.70 \\
\midrule
\multicolumn{12}{c}{\textit{Expert OCR Systems}}\\
GoogleOCR     & 85.32 & 77.46 & 63.15 & 73.40 & 64.80 & 67.67 & 67.93 & 69.72 & 57.69 & \textbf{90.62} & 71.78 \\
PP-OCRv5 MLT~\shortcite{cui2025paddleocr}      & 78.58 & 76.13 & 33.67 & 64.86 & 62.30 & 69.47 & 68.48 & 72.03 & 49.67 & 81.93 & 65.71 \\
PP-OCRv5 Det + \textbf{ScriptMoE} & \textbf{92.33} & \textbf{89.43} & 75.93 & 80.77 & \underline{81.00} & 71.81 & 72.58 & 78.41 & \textbf{79.22} & 87.45 & \textbf{80.89} \\
\bottomrule
\end{tabular}}
\end{table*}

\begin{figure}[t]
\centering
\includegraphics[width=0.55\textwidth]{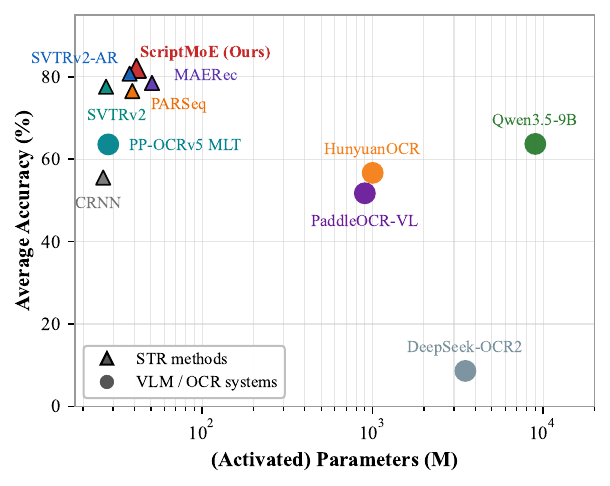}
\caption{Average accuracy on TextMuSS-Bench versus (activated) parameters. As PP-OCRv5 MLT comprises multiple models, we report the most commonly used PP-OCRv5\_server\_rec for reference.}
\label{fig:efficiency}
\end{figure}

\subsection{Main Results}

\paragraph{STR Results.} Tab.~\ref{tab:main_mlt} compares ScriptMoE against 15 representative STR models and 9 general OCR systems on TextMuSS-Bench. Among STR-class methods ScriptMoE reaches 82.06\% Avg and improves over the strongest baseline SVTRv2-AR by 1.31\%. The gains concentrate exactly where multilingual recognition is hardest, on the visually-difficult low-resource scripts Arabic (+2.98\%), Thai (+2.40\%) and Tibetan (+1.96\%). The contrast with generalist systems is far sharper: they trail by 18-70\% Avg points, while some of them collapse outright on Arabic, Bangla and Tibetan, concrete evidence that broad coverage does not imply accuracy. As Fig.~\ref{fig:efficiency} shows, ScriptMoE reaches this accuracy by activating 41.13M of its 45.85M parameters per image, one to two orders of magnitude fewer than VLM-based systems. We note two additional findings. (1) A cross-script trade-off exists: optimizing aggregate accuracy inevitably sacrifices per-script peaks. (2) The Russian gap stems from Cyrillic-Latin character confusion, compounded by Latin's dominance in real training data, a pattern our later ablations confirm.

\paragraph{End-to-End OCR Results.} These gains on STR carry over to a full OCR pipeline. Tab.~\ref{tab:ccocr} reports the end-to-end CC-OCR multilingual task with a per-language breakdown. We benchmark against three families of systems: general VLMs, OCR-specialized VLMs and expert OCR systems. Replacing the PP-OCRv5 MLT recognizer with ScriptMoE lifts the overall F1 score from 65.71\% to 80.89\%, a 15.18\% absolute jump. It also edges past the strongest zero-shot general VLM (Qwen3.5-9B, 80.73\%), the best OCR-specialized VLM (Qianfan-OCR, 76.70\%) and the expert GoogleOCR (71.78\%). This reaffirms that broad OCR pretraining does not by itself guarantee multilingual recognition accuracy. Because detection is held constant, the improvement is attributable entirely to recognition, showing that a stronger script-aware recognizer translates directly into a stronger deployed OCR system. Note that the CC-OCR task only scores text recognition, not detection boxes. Our end-to-end scores are still capped by the upstream PP-OCRv5 detector, whose missed and false detections are most visible on Latin scripts with strict word-level evaluation. Despite this, we still substantially outperform the original PP-OCRv5 and closed-source GoogleOCR.

\begin{figure}[t]
\centering
\includegraphics[width=0.72\textwidth]{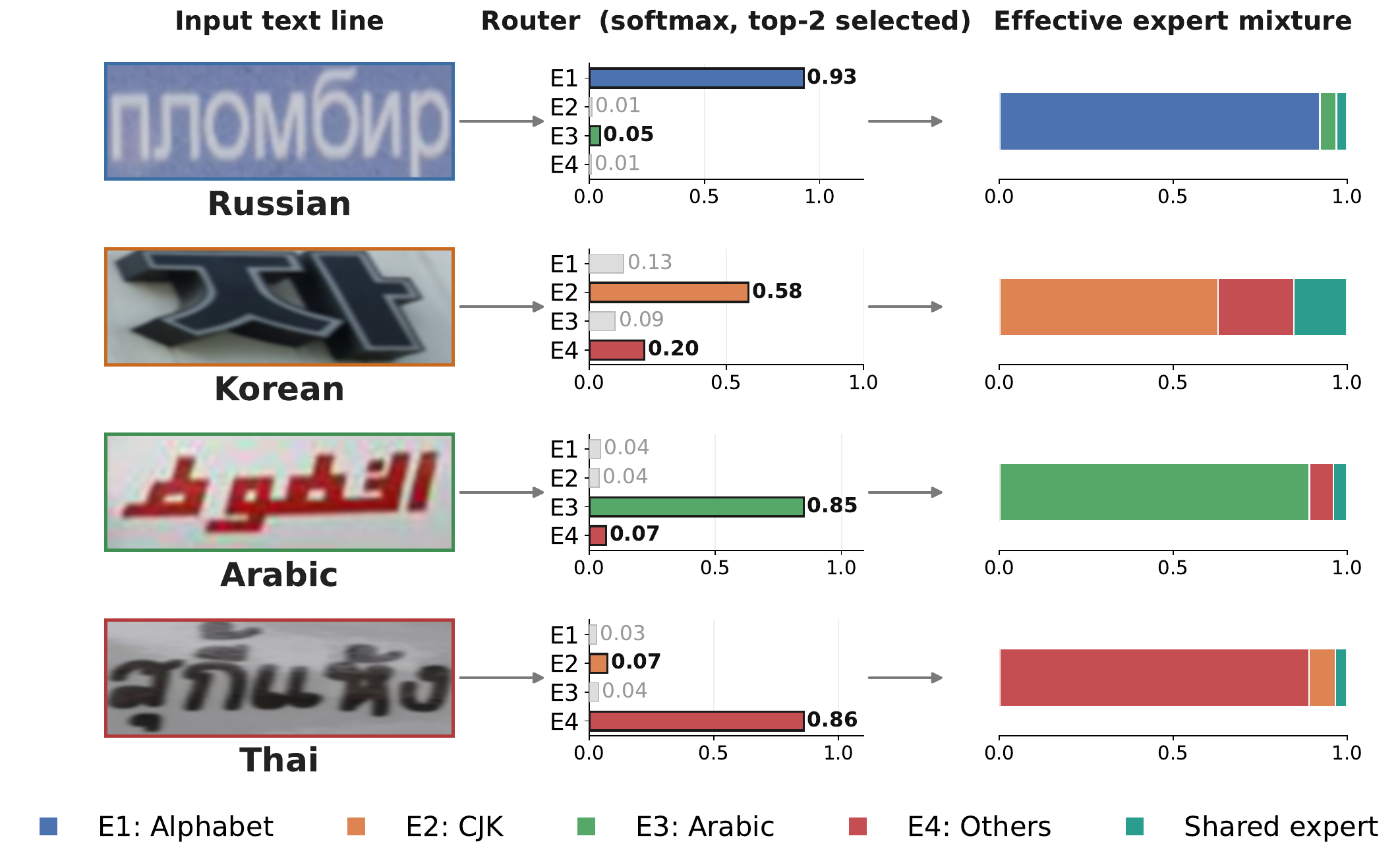}
\caption{Routing case study. One representative line per script group. Left: input, middle: router softmax over experts, right: effective mixture of the routed experts and the shared expert. Each group is dominated by a distinct expert, evidencing the learned script$\,\to\,$expert specialization.}
\label{fig:routing_example}
\end{figure}

\paragraph{Routing Analysis.}
Beyond accuracy, we verify that the router learns the intended script with expert specialization rather than an arbitrary partition. Fig.~\ref{fig:routing_example} shows, for one representative line per script group, the router softmax over the experts and the resulting effective expert mixture. As can be seen, the four lines from different scripts are each dominated by their expected expert, which strongly validates our claim.

\subsection{Ablation Study}

\begin{table*}[t]
\centering
\small
\caption{Effect of our training data. Real = all the real data mentioned, Synth = TextMuSS-10M, Synth+Real = our default. Avg is computed over the seven MLT2019 scripts.}
\label{tab:data_ablation}
\resizebox{\textwidth}{!}{%
\begin{tabular}{l|ccccccc|c|ccc}
\hline
Data & Arabic & Bangla & Chinese & Hindi & Japanese & Korean & Latin & Avg & Russian & Thai & Tibetan \\
\hline
SynthMLT~\shortcite{e2e-mlt} & 53.68 & 23.15 & 46.37 & 20.34 & 37.98 & 74.10 & 79.98 & 47.94 & 0.00    & 0.00    & 0.00    \\
Real only      & \underline{73.40} & 76.34 & \underline{91.08} & 79.39 & 58.25 & 68.92 & \textbf{92.47} & 77.12 & 0.00    & 0.00    & 0.00    \\
Synth only     & 60.00 & \underline{83.21} & \underline{91.08} & \underline{82.44} & \underline{67.85} & \underline{85.71} & 86.95 & \underline{79.61} & \textbf{68.13} & \underline{70.30} & \underline{87.08} \\
Synth + Real   & \textbf{78.09} & \textbf{88.30} & \textbf{95.38} & \textbf{86.77} & \textbf{71.21} & \textbf{87.19} & \underline{91.67} & \textbf{85.52} & \underline{61.20} & \textbf{72.00} & \textbf{88.76} \\
\hline
\end{tabular}}
\end{table*}

\begin{table*}[t]
\centering
\small
\caption{Full ablation on TextMuSS-Bench. The experimental setup is identical to the main experiment. \#Experts$=$0 denotes the pure AR baseline without MoE.}
\label{tab:ablation_full}
\resizebox{\textwidth}{!}{%
\begin{tabular}{ccccc|cccccccccc|c}
\hline
\#Experts & Top-experts & Routing-level & $\mathcal{L}_{\text{scls}}$ & Shared expert & Arabic & Bangla & Chinese & Hindi & Japanese & Korean & Latin & Russian & Thai & Tibetan & Avg \\
\hline
0 & / & / & / & / & 75.11 & 87.79 & 94.15 & 87.28 & 69.36 & 85.86 & \textbf{92.27} & 59.30 & 69.60 & 86.80 & 80.75 \\
2 & 2 & image & \checkmark & \checkmark & \textbf{80.85} & 87.53 & 95.08 & 86.26 & 69.53 & 87.19 & 91.72 & \textbf{63.47} & 70.67 & 82.87 & 81.52 \\
4 & 2 & image & \checkmark & \checkmark & 78.09 & 88.30 & 95.38 & 86.77 & \textbf{71.21} & 87.19 & 91.67 & 61.20 & \textbf{72.00} & \textbf{88.76} & \textbf{82.06} \\
10 & 2 & image & \checkmark & \checkmark & 76.81 & 88.04 & 93.23 & 87.28 & 70.37 & 87.48 & 91.81 & 58.25 & \textbf{72.00} & 87.92 & 81.32 \\
4 & 1 & image & \checkmark & \checkmark & 78.30 & 88.55 & 94.15 & 87.28 & 70.54 & 87.48 & 91.47 & 60.53 & 69.47 & 88.20 & 81.60 \\
4 & 4 & image & \checkmark & \checkmark & 79.15 & 88.55 & \textbf{95.69} & 87.53 & 68.69 & 86.89 & 92.03 & 62.14 & 71.33 & 87.36 & 81.94 \\
4 & 2 & token & \checkmark & \checkmark & 78.09 & \textbf{88.80} & \textbf{95.69} & \textbf{89.06} & 68.69 & \textbf{87.92} & 92.10 & 59.58 & 69.60 & 87.36 & 81.69 \\
4 & 2 & image &  & \checkmark & 77.66 &	87.79	& 95.08 &	85.75	&70.88	&87.33	&91.93	&61.67	&70.00	&88.20	&81.63  \\
4 & 2 & image & \checkmark &  &  74.47	&88.30	&95.08	&87.28	&70.03	&87.04	&91.55	&63.19	&71.47	&85.11	&81.35  \\
\hline
\end{tabular}}
\end{table*}

\paragraph{Data Ablation}

Tab.~\ref{tab:data_ablation} disentangles the role of real and synthetic data. First, real-only training collapses on the three low-resource scripts that it never sees. It confirms that synthetic supplementation is mandatory for at least these scripts. Training on TextMuSS-10M alone already far surpasses SynthMLT and even outperforms real data on the MLT2019 average by +2.49\%, demonstrating the benefit of sheer synthetic scale. More practically, combining synthetic and real data yields an effective complement: the MLT2019 average rises by 8.40\% over real-only training. An interesting pattern emerges in the per-script breakdown: adding synthetic data lowers Latin, while adding real data lowers Russian. This is primarily because Latin and Cyrillic share visually identical glyphs that map to different characters, causing cross-script confusion. ScriptMoE mitigates this confusion relative to other STR methods, but the gap remains difficult to close entirely because Latin benefits from abundant real data whereas Cyrillic relies solely on synthetic substitutes.

\paragraph{Model Ablation}
Tab.~\ref{tab:ablation_full} isolates the key components of the MoE decoder and three trends emerge. The MoE decoder is indispensable: the pure AR baseline without experts falls behind MoE series except Latin, the single most data-rich script, confirming that a single dense decoder cannot adequately serve all ten scripts. Four experts reaches the best average of 82.06\%, with the gain concentrated on the scripts that benefit most from dedicated capacity (Thai +2.40\%, Chinese +1.23\%). Scaling to ten experts (one per script) however, erases the gain: each expert now sees too few samples to specialize, and some scripts are better served by sharing an expert with their structural neighbors (e.g.\ Chinese, Japanese and Korean share stroke topology). Top-2 is the right sparsity. Top-1 still outperforms the no-MoE baseline, but a single routed expert cannot capture the shared sub-patterns between related scripts and drops most on Thai and Chinese. Top-4 nearly matches Top-2 on average yet activates more parameters per image for a negligible return, and actually hurts Japanese and Korean, indicating diminishing returns from over-activation. Routing granularity is essentially a wash: token-level routing trades marginal per-script gains on some scripts (e.g.\ Hindi) for comparable losses on others (e.g.\ Russian), all within noise. We therefore keep image-level routing for its lower router cost. Finally, the last two rows isolate our two script-aware components. Removing the script-classification signal lets the router drift away from script boundaries and costs 0.43\% on average. Removing the shared expert is more damaging (-0.71\% on average), with the drop concentrated on the low-resource scripts: Arabic (-3.62\%) and Tibetan (-3.65\%).

\section{Conclusion}
In this paper, we have presented a systematic study of all-in-one multilingual scene text recognition from both data and model perspectives. On the data side, we constructed TextMuSS-10M, a large-scale multilingual synthetic scene text dataset covering 10 scripts and 229 languages, which provides balanced and sufficient training signal for scripts where real data is scarce. We also assembled TextMuSS-Bench, a real scene text benchmark spanning all ten scripts for comprehensive evaluation. On the model side, we proposed ScriptMoE, a script-aware Mixture-of-Experts architecture that keeps a shared visual encoder and reallocates decoder capacity through an image-level router activating script-aligned experts. With the shared expert preserving cross-script transfer and a lightweight script-classification signal guiding expert specialization, ScriptMoE maintains the efficiency of a single model while effectively serving diverse scripts. Extensive experiments demonstrate ScriptMoE's effectiveness: it achieves competitive average accuracy on TextMuSS-Bench, with notable gains on low-resource scripts such as Arabic, Thai, and Tibetan. When integrated into an end-to-end OCR pipeline, it surpasses both expert OCR systems and strong VLMs on the CC-OCR multilingual task while remaining lightweight. In the future, we plan to explore continual learning strategies that allow new scripts to be added without retraining the full model. We are also interested in integrating ScriptMoE with stronger text detectors to further improve end-to-end multilingual OCR performance.

\bibliographystyle{plainnat}
\bibliography{main}

\clearpage

\newpage

\section*{Appendix}

\subsection*{Dataset and Benchmark Details}
\label{sec:supp_data}

\subsubsection*{Data Card of TextMuSS-10M}
\label{sec:supp_data10m}

\paragraph{Text-source composition.}
For every script, the rendered text strings are drawn from four sources with a fixed mixing ratio: $40\%$ real words sampled from the per-language lexicon, $20\%$ words with their characters randomly shuffled (to expose the model to out-of-vocabulary and non-linguistic character sequences), $20\%$ vocabulary-expansion strings that oversample rare characters so that the full character table is covered, and $20\%$ sentences sampled from the News Crawl corpora. This mixture balances lexical realism with full coverage of the character space. For the Latin script, which internally spans many languages, the lexicon is sampled across all Latin-based languages to keep the intra-script language distribution balanced.

\paragraph{Rendering effects.}
Unless a template specifies otherwise, each rendering effect is applied independently with the following probability and may be stacked with the others: $20\%$ curved layout, $20\%$ multi-directional (including vertical) layout, and $20\%$ perspective distortion. Backgrounds are sampled from the 8k background images released by SynthText~\cite{st}, a widely used and publicly available resource for STR data synthesis.

\begin{table}[ht]
\centering
\small
\caption{Per-script character-table sizes.}
\label{tab:supp_charset}
\begin{tabular}{l*{10}{c}}
\toprule
Script  & Arabic & Bangla & Chinese & Hindi & Japanese & Korean & Latin & Cyrillic & Thai & Tibetan \\
\midrule
\#Chars & 747    & 102    & 16{,}147 & 817   & 4{,}398  & 3{,}687 & 923   & 850      & 524  & 64      \\
\bottomrule
\end{tabular}
\end{table}

\paragraph{Character set.}
Tab.~\ref{tab:supp_charset} lists the size of each per-script character table. We build the released character set by de-duplicating and merging all ten tables into a single unified vocabulary and adding the space symbol, yielding 19,684 characters. Together with the three special tokens (BOS/EOS/PAD), the decoder embedding uses a vocabulary of 19,687 entries.

\subsubsection*{Collection and Annotation of TextMuSS-Bench}
\label{sec:supp_benchdata}

TextMuSS-Bench extends the MLT2019~\cite{nayef2019icdar2019} test split with three additional scripts (Russian, Thai and Tibetan) that we collect and annotate ourselves. The collection and annotation pipeline is as follows.

\paragraph{Image collection.}
For each of the three new scripts, we collect real-world scene images of the corresponding language from publicly available sources.

\paragraph{Detection and box verification.}
We first run an OCR text detector (PP-OCRv5~\cite{cui2025paddleocr}) to obtain candidate text regions, and then manually verify and correct the detected bounding boxes so that every box tightly encloses a real text instance.

\paragraph{Transcription annotation.}
Text belonging to each target script is transcribed by a language expert of that script. Because experts in these low-resource languages are scarce, each script is annotated by a single expert.

\paragraph{Quality control.}
All annotations are then subjected to a unified quality check by a separate reviewer (not specialized in these low-resource languages), who verifies at the morphological level that the transcribed text visually matches the text in the image.

\begin{table*}[htbp]
\centering
\small
\caption{Full language-to-script mapping. The ten scripts are organized into the four expert families used by the router. All languages representable by our released character set are enumerated per script. This set can represent 229 languages. Languages written in more than one script (e.g.\ Kurdish, in both Latin and Arabic) are listed under each script they appear in. Greek uses its own alphabet, but because several Greek letters coincide with characters already shared by Latin-script text, we fold it into the Latin (alphabet) group for convenience rather than treating it as a separate script.}
\label{tab:supp_lang_groups}
\begin{tabular}{l l p{0.66\textwidth}}
\toprule
Expert group & Script & Languages \\
\midrule
Alphabet & Latin & English, French, German, Afrikaans, Italian, Spanish, Bosnian, Portuguese, Czech, Welsh, Danish, Estonian, Irish, Croatian, Uzbek, Hungarian, Serbian-Latin, Indonesian, Occitan, Icelandic, Lithuanian, Maori, Malay, Dutch, Norwegian, Polish, Slovak, Slovenian, Albanian, Swedish, Swahili, Tagalog, Turkish, Somali, Azerbaijani, Kurdish-Latin, Latvian, Maltese, Zulu, Romanian, Vietnamese, Finnish, Basque, Galician, Luxembourgish, Romansh, Catalan, Quechua, Xhosa, Sesotho, Shona, Kinyarwanda, Rundi, Ndebele, Swati, Bemba, Tumbuka, Kongo, Tsonga, Oromo, Cebuano, Hiligaynon, Ilocano, Waray, Bikol, Pangasinan, Bislama, Tok Pisin, Tetum, Fijian, Kikuyu, Ganda, Chichewa, Tswana, Sango, Wolof, Malagasy, Turkmen, Faroese, Frisian, Corsican, Sardinian, Breton, Walloon, Aragonese, Asturian, Scottish Gaelic, Manx, Cornish, Kashubian, Sorbian, Silesian, Haitian Creole, Aymara, Nahuatl, Guarani, Papiamento, Sranan, Greenlandic, Zaza, Ladino, Chamorro, Greek \\
\addlinespace
 & Cyrillic & Russian, Belarusian, Ukrainian, Serbian-Cyrillic, Bulgarian, Mongolian, Abkhaz, Adyghe, Kabardian, Avar, Dargwa, Ingush, Chechen, Lak, Lezgian, Tabasaran, Kazakh, Kyrgyz, Tajik, Macedonian, Tatar, Chuvash, Bashkir, Mari, Mordvin, Moldovan, Udmurt, Komi, Ossetian, Buryat, Kalmyk, Tuvan, Sakha, Karakalpak, Crimean Tatar, Rusyn, Kumyk, Karachay-Balkar, Abaza, Nogai, Khakas, Even, Evenki, Shor, Rutul, Tsakhur, Aghul, Komi-Permyak, Erzya, Moksha, Khanty, Mansi, Nenets, Chukchi, Koryak, Dungan, Tofa, Yukaghir, Nanai, Selkup \\
\midrule
CJK & Chinese & Simplified Chinese, Traditional Chinese \\
 & Japanese & Japanese \\
 & Korean & Korean \\
\midrule
Arabic & Arabic & Arabic, Persian, Uyghur, Urdu, Pashto, Kurdish-Arabic, Sindhi, Balochi, Saraiki, Punjabi, Brahui, Kashmiri, Luri, Mazandarani, Gilaki, Wakhi, Shughni, Rohingya \\
\midrule
Others & Hindi & Hindi, Marathi, Nepali, Awadhi, Maithili, Angika, Bhojpuri, Magahi, Nagpuri, Santali, Newari, Konkani, Sanskrit, Haryanvi, Chhattisgarhi, Rajasthani, Bundeli, Braj, Garhwali, Kumaoni, Dogri, Bodo, Sherpa, Magar, Kurukh, Ho, Mundari, Sadri, Kanauji, Marwari, Mewari \\
 & Bangla & Bangla, Bishnupriya, Sylheti, Chakma, Rangpuri, Meitei, Kokborok \\
 & Thai & Thai, Isan, Southern Thai \\
 & Tibetan & Tibetan, Dzongkha, Ladakhi, Sikkimese, Balti \\
\midrule
\multicolumn{2}{l}{\textbf{Total}} & \textbf{229 languages across 10 scripts} \\
\bottomrule
\end{tabular}
\end{table*}

\subsection*{Full Language-to-Script Mapping}
\label{sec:supp_lang}

We consolidate the major world languages into ten representative scripts, and these ten scripts can represent 229 languages that share this common character space. For modeling, the ten scripts are further organized into the four expert groups used by the router: the alphabet group, the CJK group, the Arabic family, and the Others group. Tab~\ref{tab:supp_lang_groups} gives the complete per-language enumeration for each script. All of these languages are representable by our released character set (19,684 characters including the space symbol).

\subsection*{Implementation Details of ScriptMoE}
\label{sec:supp_method}

The main paper defines ScriptMoE at the formulation level (see Eq.~(1)--(4) of the main paper). Here we complement it with the exact network configuration and training objective as implemented, so that the model can be reproduced without ambiguity. Unless stated otherwise, the values below are the defaults used in all experiments.

\subsubsection*{Network Configuration}

\paragraph{Visual encoder.}
We use the hierarchical SVTRv2 encoder~\cite{du2024svtrv2} shared across all scripts. It has three stages with widths $[128,256,384]$, depths $[6,6,6]$ and $[4,8,12]$ attention heads. The first stage uses only $5\times5$ local convolutional mixers. The second stage combines convolutional blocks with global self-attention and applies a $[2,1]$ vertical sub-sampling, and the third stage is fully global. Positional embeddings are disabled, and the encoder emits a sequence of $d$-dimensional visual tokens $\mathbf{F}$ that feed both the router and the decoder.

\paragraph{Script-aware decoder.}
The decoder stacks 2 Transformer blocks with 12 attention heads. We set the number of decoder blocks to $N{=}2$ to align with the standard SVTRv2-AR~\cite{ye2026wrong} configuration, so that ScriptMoE and its autoregressive baseline differ only in the MoE-FFN rather than in decoder depth. Each block applies self-attention, cross-attention, and then the script-aware MoE-FFN, each wrapped in a post-norm residual connection. Target tokens are embedded (scaled by $\sqrt{d}$) with sinusoidal positional encoding, and a linear head projects the final states to the character vocabulary. Decoding is greedy autoregressive with dedicated BOS/EOS symbols. The maximum length is 25 for the word-level STR setting and 100 for the line-level end-to-end variant. We illustrate the forward pass of a script-aware MoE-FFN layer (full procedure in Alg.~\ref{alg:supp_moe}).

\begin{algorithm}[t]
\caption{Forward pass of a script-aware MoE-FFN layer}
\label{alg:supp_moe}
\begin{algorithmic}[1]
\STATE \textbf{Input:} decoder states $H\in\mathbb{R}^{B\times T\times d}$, visual memory $F\in\mathbb{R}^{B\times L\times d}$
\STATE $\bar{F}_b \leftarrow \mathrm{mean}_i F_{b,i}$ \hfill// sample-level router input
\STATE \textbf{if} training: $\bar{F}_b \leftarrow \bar{F}_b\odot(1+\sigma\,\epsilon_b)$, \; $\epsilon_b\sim\mathcal{N}(0,I)$
\STATE $\mathbf{p}_b \leftarrow \mathrm{softmax}(W_g\bar{F}_b)$, \; $\mathcal{K}_b \leftarrow \mathrm{Top\text{-}2}(\mathbf{p}_b)$
\STATE $g_{b,e} \leftarrow p_{b,e}\big/\!\sum_{e'\in\mathcal{K}_b}p_{b,e'}$ \; for $e\in\mathcal{K}_b$ \hfill// renormalize
\STATE $y^{\text{rt}}_b \leftarrow \sum_{e\in\mathcal{K}_b} g_{b,e}\, f_e(H_b)$ \hfill// shared by all $T$ tokens
\STATE $y^{\text{sh}}_b \leftarrow f_{\text{share}}(H_b)$
\STATE $\alpha_b \leftarrow \mathrm{sigmoid}(\mathbf{w}_s^{\top}H_b)$ \hfill// learnable per-token gate
\STATE \textbf{return} $\alpha_b\, y^{\text{sh}}_b + (1-\alpha_b)\, y^{\text{rt}}_b$
\end{algorithmic}
\end{algorithm}

\subsubsection*{Main Training Objective Details}

In Eq.~(5) of the main paper, $\mathcal{L}_{\text{ar}}$ is the token-level autoregressive cross-entropy over the target transcription $\mathbf{y}=(y_1,\dots,y_T)$ conditioned on the visual memory $\mathbf{F}$,
\begin{equation}
\mathcal{L}_{\text{ar}} = -\frac{1}{T}\sum_{t=1}^{T}\sum_{c\in\mathcal{V}} \tilde{y}_{t,c}\,\log p\!\left(c \mid y_{<t}, \mathbf{F}\right),
\end{equation}
where $\mathcal{V}$ is the character vocabulary and $\tilde{y}_{t,c}$ is the label-smoothed target distribution with smoothing $\varepsilon{=}0.1$.

\subsubsection*{Parameter and Efficiency Breakdown}
\label{sec:supp_params}

Tab.~\ref{tab:supp_param_breakdown} decomposes the parameters of ScriptMoE. The full model stores 45.85M parameters. Because routing is performed at the sample level with Top-2 selection over 4 experts, each forward pass activates only 2 of the 4 experts per layer, so on average 41.13M parameters (89.69\%) are used per sample and the remaining 4.73M sit idle as unrouted experts. Relative to a dense-FFN decoder (the 4 experts collapsed into a single equivalent FFN), the MoE design stores about $3\times1.18{=}3.5$M extra ``stored-but-inactive'' parameters, while the activated compute stays comparable to a slightly larger FFN (Top-2 experts plus one shared expert). The 9.45M of expert parameters (2 layers $\times$ 4 experts) is where the script specialization is concentrated.

\begin{table}[t]
\centering
\small
\caption{Parameter breakdown of ScriptMoE.}
\label{tab:supp_param_breakdown}
\begin{tabular}{l r}
\toprule
Component & Params \\
\midrule
\multicolumn{2}{l}{\textit{Overview}} \\
Total (stored) & 45.85\,M \\
Activated (per forward) & 41.13\,M \\
Inactive experts (skipped by router) & 4.73\,M \\
Activation ratio & 89.69\% \\
\midrule
\multicolumn{2}{l}{\textit{Module decomposition}} \\
SVTRv2 Encoder & 17.65\,M \\
Decoder (ScriptMoE, all) & 28.20\,M \\
\quad Token embedding & 7.56\,M \\
\quad Output projection & 7.56\,M \\
\quad Script classifier & 0.07\,M \\
\quad $2\times$ MoE decoder layer & 13.01\,M \\
\midrule
\multicolumn{2}{l}{\textit{Single MoE decoder layer ($\times2$)}} \\
Self-Attn $+$ Cross-Attn $+$ $3\times$LN & 1.19\,M \\
Router & 0.0015\,M \\
Shared expert & 0.59\,M \\
Single expert FFN & 1.18\,M \\
All 4 experts & 4.73\,M \\
Per-layer stored total & 6.50\,M \\
Per-layer Top-2 activated & 4.14\,M \\
\bottomrule
\end{tabular}
\end{table}

\begin{table}[t]
\centering
\small
\caption{Inference efficiency of ScriptMoE and representative baselines, measured on a single NVIDIA V100 with batch size 256.}
\label{tab:supp_efficiency}
\begin{tabular}{l r r r r}
\toprule
Method & Params\,(M) & Latency\,(ms) & Throughput\,(img/s) & Mem\,(MB) \\
\midrule
CRNN~\shortcite{shi2017crnn}             & 26.25 & 57.40  & 4459.9 & 1368.4 \\
SVTRv2~\shortcite{du2024svtrv2}           & 27.30 & 210.32 & 1217.2 & 1560.8 \\
PARSeq~\shortcite{parseq}        & 38.90 & 260.17 & 984.0  & 1254.0 \\
MAERec~\shortcite{jiang2023revisiting}   & 50.73 & 1297.63 & 197.3 & 2411.3 \\
SVTRv2-AR~\shortcite{ye2026wrong}       & 37.50 & 395.92 & 646.6  & 2057.3 \\
\textbf{ScriptMoE} & 45.85 & 541.07 & 473.1 & 2091.0 \\
\bottomrule
\end{tabular}
\end{table}

Tab.~\ref{tab:supp_efficiency} reports inference latency, throughput and peak memory against representative STR baselines under an identical batch size of 256 on a single V100. Among autoregressive models, ScriptMoE stays well below the heavy MAERec in both latency and memory, while its sample-level MoE keeps the activated cost close to the standard AR decoder. In terms of training cost, the full model is trained for 2 epochs on 8 V100 GPUs, taking about 92.7 GPU-hours (11.8 hours wall-clock).

\begin{table*}[t]
\centering
\small
\caption{Reference benchmarks: BCTR~\cite{bctr} (Chinese) and Union14M-Benchmark~\cite{jiang2023revisiting} (English). Best in \textbf{bold}, second \underline{underlined}.}
\label{tab:supp_ref}
\resizebox{\textwidth}{!}{%
\begin{tabular}{l|cccc|c||ccccccc|c}
\toprule
& \multicolumn{5}{c||}{BCTR} & \multicolumn{8}{c}{Union14M-Benchmark} \\
Method & Document & Handwriting & Scene & Web & Avg & Curve & Multi-Oriented & Artistic & Contextless & Salient & Multi-Words & General & Avg \\
\midrule
SVTRv2~\shortcite{du2024svtrv2}   & 99.49 & 70.13 & 78.48 & 87.28 & 83.85 & 91.22 & 94.67 & 78.11 & 85.37 & 85.99 & 88.30 & 82.05 & 86.53 \\
ABINet~\shortcite{abinet}   & 96.92 & 51.56 & 67.31 & 80.24 & 74.01 & 76.75 & 74.51 & 66.22 & 73.43 & 72.43 & 62.00 & 74.84 & 71.45 \\
PARSeq~\shortcite{parseq}   & 98.58 & 65.70 & 78.08 & 86.77 & 82.28 & 81.53 & 89.55 & 71.44 & 82.03 & 80.38 & 84.20 & 81.40 & 81.51 \\
MAERec~\shortcite{jiang2023revisiting}   & 98.98 & 70.78 & 81.72 & 88.71 & 85.05 & 90.48 & 93.64 & 80.22 & 87.42 & \underline{88.71} & 89.51 & \textbf{84.43} & 87.77 \\
CDistNet~\shortcite{zheng2024cdistnet} & 97.94 & 65.87 & 76.44 & 85.97 & 81.56 & 88.09 & 90.80 & 74.44 & 84.08 & 84.61 & 87.09 & 81.54 & 84.38 \\
SMTR~\shortcite{smtr}     & 99.14 & 66.62 & 78.07 & 87.48 & 82.83 & 89.82 & 94.59 & 77.67 & 84.72 & 85.36 & 87.82 & 82.41 & 86.06 \\
SVTRv2-AR~\shortcite{ye2026wrong}& \underline{99.35} & \underline{73.12} & \underline{81.99} & \underline{89.25} & \underline{85.93} & \underline{93.12} & \underline{96.49} & \underline{80.22} & \textbf{88.06} & 88.68 & \textbf{90.35} & 83.78 & \underline{88.67} \\
\textbf{\ours} & \textbf{99.47} & \textbf{74.79} & \textbf{83.35} & \textbf{89.80} & \textbf{86.85} & \textbf{93.49} & \textbf{96.71} & \textbf{81.78} & \underline{87.55} & \textbf{89.10} & \underline{89.87} & \underline{84.15} & \textbf{88.95} \\
\bottomrule
\end{tabular}}
\end{table*}

\subsection*{Additional Implementation Details}
\label{sec:supp_impl}

\paragraph{Training Details.}
All models are trained with the OpenOCR framework on 8 NVIDIA V100 GPUs (32\,GB each) and an Intel Xeon Platinum 8255C CPU, under a Linux environment. For a fair comparison, all 15 STR baselines are re-trained under the identical data, schedule and hardware, and re-evaluated on TextMuSS-Bench, using the default model-specific configuration (e.g., network depth, hidden dimension) provided by the OpenOCR framework for each architecture. We follow the training recipe of SVTRv2~\cite{du2024svtrv2}. We use the AdamW~\cite{adamw} optimizer with a weight decay of 0.05. The peak learning rate is set to $6.5\times10^{-4}$ and the global batch size is 1024 (8 GPUs $\times$ 128 per GPU). A one-cycle learning-rate scheduler with a 1.5-epoch linear warm-up is used across the total of 2 training epochs. We fix a global random seed for all experiments. The only remaining source of stochasticity is the router jitter $\epsilon\sim\mathcal{N}(0,I)$ in Eq.~(3) of the main paper, which is likewise seeded so that runs are reproducible. Following PARSeq~\cite{parseq}, we randomly apply rotation, perspective distortion, motion blur and Gaussian noise during training. All experiments are run with Python 3.8, PyTorch 2.2.0, CUDA 11.8.

\paragraph{Task Adaptation Details.}
For the STR experiments, the maximum text length is set to 25 during training. Because the PP-OCRv5 detector operates at the line level rather than the word level, we extend the maximum length to 100 and re-train ScriptMoE to handle line-level inputs for the end-to-end variant. At inference, we set the detection threshold to 0.0 and the recognition threshold to 0.7 to produce the final end-to-end OCR results.

\paragraph{Evaluation Details.}
We report word accuracy as the primary STR metric, following the mainstream evaluation convention in the STR literature. Unlike character-level metrics such as CER or edit distance, word accuracy requires the whole predicted string to exactly match the ground truth. For scene text recognition, a meaningful and usable result is a correctly recognized text instance as a whole. A single-character error already renders the output wrong for many downstream applications. We therefore adopt this stricter, application-aligned criterion. Following the common English evaluation protocol, we first lowercase both the prediction and the ground truth, then remove Unicode symbols and whitespace before computing both metrics, so that the scores reflect accuracy on meaningful text content. As a complementary character-level measure, we additionally report $1-\mathrm{NED}$, where NED is the normalized edit distance between the prediction and ground truth. Higher $1-\mathrm{NED}$ indicates a smaller character-level discrepancy and thus better recognition quality.

\subsection*{VLM Evaluation Protocol}
\label{sec:supp_protocol}

\paragraph{Prompts on TextMuSS-Bench.}
For general-purpose VLMs we use a single unified instruction across all scripts:
\begin{AIbox}{General VLM Prompt:}
``Please directly output all original readable text in the image. Do not include any explanation or description. Only output the recognized text.''
\end{AIbox}
For OCR-specialized VLMs we adopt each model's own default OCR prompt rather than a shared instruction:
\begin{AIbox}{DeepSeek-OCR2~\cite{wei2026deepseek} Prompt:}
``<image>\textbackslash nFree OCR.''
\end{AIbox}
\begin{AIbox}{PaddleOCR-VL~\cite{cui2025paddleocrvl} Prompt:}
``OCR:''
\end{AIbox}
\begin{AIbox}{GOT-OCR2.0~\cite{wei2024general} Prompt:}
``ocr''
\end{AIbox}
\begin{AIbox}{HunyuanOCR~\cite{team2025hunyuanocr} Prompt:}
``Extract the text from this image.''
\end{AIbox}
\begin{AIbox}{GLM-OCR~\cite{duan2026glm} Prompt:}
``Text Recognition:''
\end{AIbox}

\paragraph{Output normalization.}
The cleaning/alignment rules are identical to those used for our own model as described in Evaluation Details. Predictions and ground truth are lowercased, and Unicode symbols and whitespace are removed before scoring. The same normalization is applied to the VLM outputs so that all systems are compared under one protocol. 

\begin{figure}[h]
    \centering
    \includegraphics[width=0.8\linewidth]{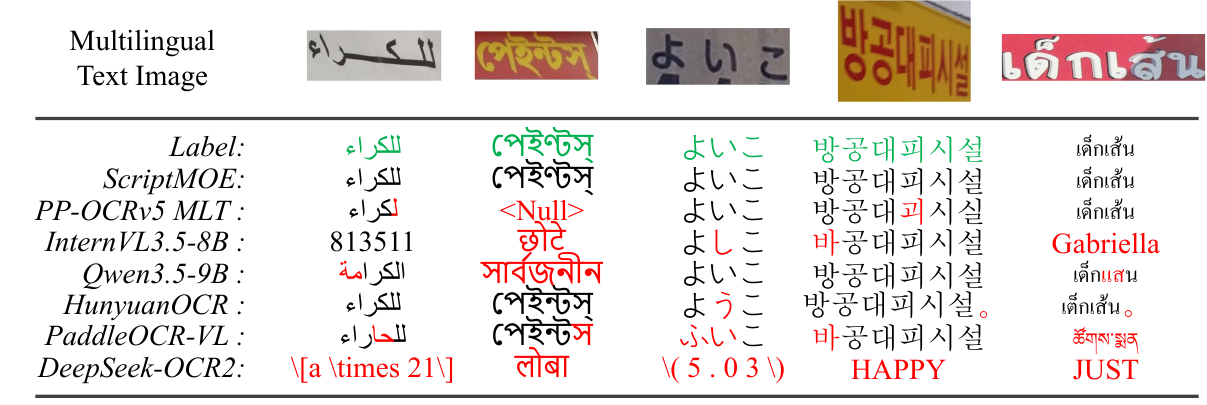}
    \caption{Visualization of different models' predictions on TextMuSS-Bench. 
    Characters differing from the ground truth are highlighted in {\color{red}red}, 
    and \texttt{<Null>} indicates that the model does not support the corresponding language.}
    \label{fig:vis_result}

\end{figure}

\paragraph{Qualitative comparison.}
Fig.~\ref{fig:vis_result} compares the predictions of ScriptMoE against several representative VLMs and multilingual OCR systems on TextMuSS-Bench. ScriptMoE consistently produces correct results, while competing methods often fail to guarantee accurate multilingual recognition in general.

\begin{figure}[h]
    \centering
    \includegraphics[width=0.8\linewidth]{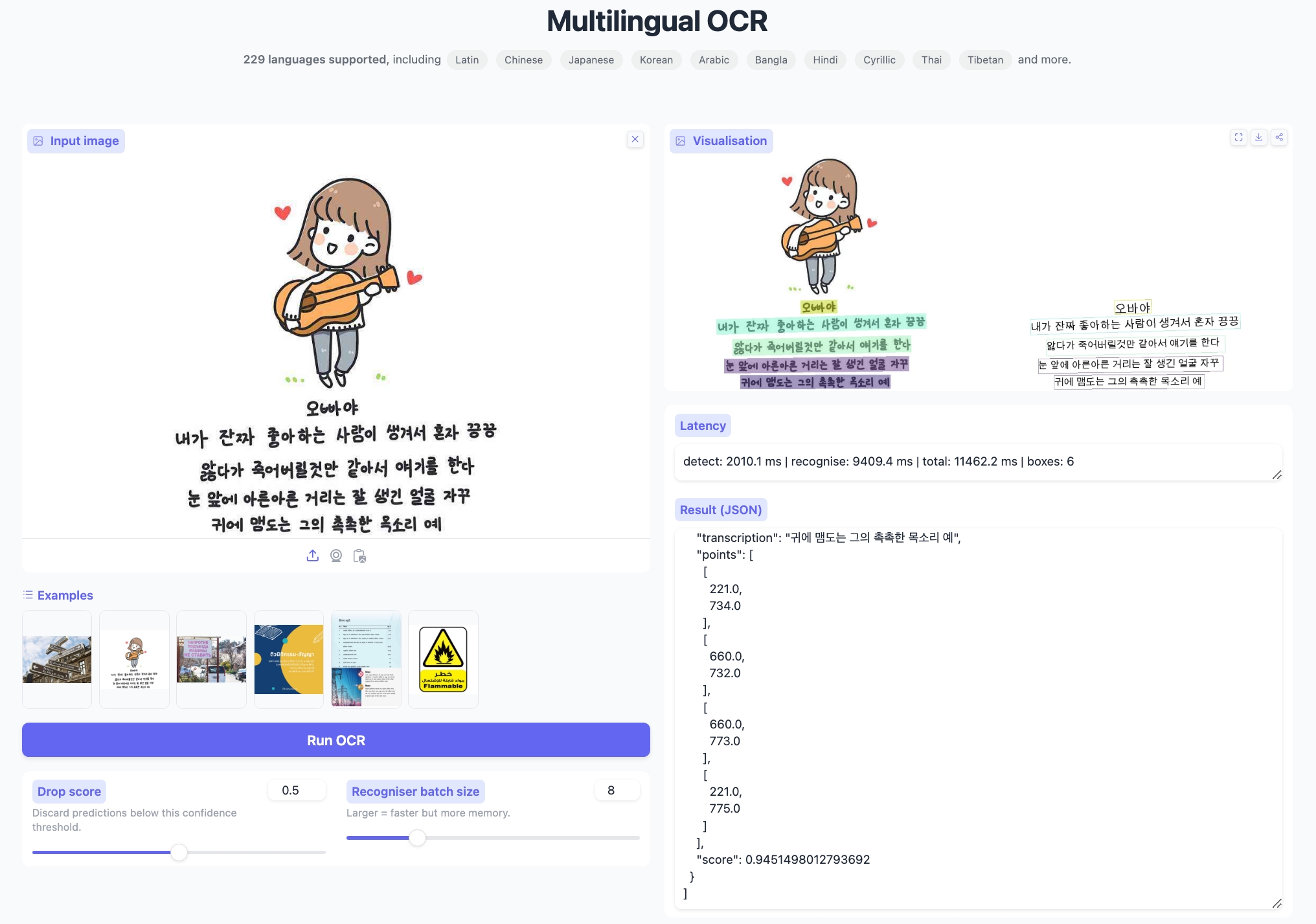}
    \caption{Users will upload a multilingual scene text image, select a confidence threshold for filtering, and click ``Run OCR'' to obtain the end-to-end OCR recognition results.}
    \label{fig:demo}
\end{figure}

\paragraph{CC-OCR protocol.}
For the end-to-end CC-OCR~\cite{ccocr} evaluation, we strictly follow its official evaluation protocol and metrics: Latin-script languages are scored at the word level, while all other languages are scored at the character level. To complement the quantitative evaluation, we additionally provide an end-to-end OCR demo, as illustrated in Fig.~\ref{fig:demo}.

\begin{table*}[t]
\centering
\small
\caption{Extended ablations on TextMuSS-Bench. Each block varies a single hyper-parameter around the shared default configuration. Best Avg within each block in \textbf{bold}.}
\label{tab:supp_ablation}
\resizebox{\textwidth}{!}{%
\begin{tabular}{l|cccccccccc|c}
\toprule
Setting & Arabic & Bangla & Chinese & Hindi & Japanese & Korean & Latin & Russian & Thai & Tibetan & Avg \\
\midrule
\multicolumn{12}{l}{\textit{Script-classification loss weight} $\lambda_{\text{scls}}$} \\
0.0          & 77.66 & 87.79 & 95.08 & 85.75 & 70.88 & 87.33 & 91.93 & 61.67 & 70.00 & 88.20 & 81.63 \\
0.1 (default) & 78.09 & 88.30 & 95.38 & 86.77 & 71.21 & 87.19 & 91.67 & 61.20 & 72.00 & 88.76 & \textbf{82.06} \\
0.2          & 76.60 & 88.55 & 94.46 & 87.28 & 70.71 & 87.19 & 91.62 & 61.29 & 68.53 & 88.20 & 81.44 \\
0.5          & 78.09 & 87.28 & 94.46 & 86.77 & 71.72 & 87.19 & 91.50 & 61.20 & 70.80 & 84.55 & 81.35 \\
\midrule
\multicolumn{12}{l}{\textit{Shared-expert width ratio}} \\
0.0          & 74.47 & 88.30 & 95.08 & 87.28 & 70.03 & 87.04 & 91.55 & 63.19 & 71.47 & 85.11 & 81.35 \\
0.5 (default) & 78.09 & 88.30 & 95.38 & 86.77 & 71.21 & 87.19 & 91.67 & 61.20 & 72.00 & 88.76 & \textbf{82.06} \\
1.0          & 78.94 & 88.55 & 95.08 & 86.26 & 70.37 & 87.78 & 91.79 & 59.87 & 70.67 & 84.83 & 81.41 \\
\midrule
\multicolumn{12}{l}{\textit{Router jitter} $\sigma$} \\
0.0          & 78.51 & 88.55 & 95.38 & 87.28 & 71.04 & 87.63 & 91.72 & 62.05 & 69.33 & 84.55 & 81.61 \\
0.05 (default) & 78.09 & 88.30 & 95.38 & 86.77 & 71.21 & 87.19 & 91.67 & 61.20 & 72.00 & 88.76 & \textbf{82.06} \\
0.1          & 77.87 & 87.53 & 94.46 & 87.28 & 70.88 & 87.48 & 91.71 & 61.29 & 69.87 & 84.83 & 81.32 \\
\bottomrule
\end{tabular}}
\end{table*}

\subsection*{Monolingual Reference Benchmarks}

To verify that script-aware specialization does not siphon capacity away from the dominant scripts, Tab.~\ref{tab:supp_ref} reports results on the long-saturated Benchmarking Chinese Text Recognition~\cite{bctr} (BCTR) benchmark and the English Union14M-Benchmark~\cite{jiang2023revisiting}. ScriptMoE remains competitive on both, reaching 86.85\% Avg on BCTR (+0.92\% over SVTRv2-AR) and 88.95\% Avg on Union14M-Benchmark (+0.28\%), confirming that the multilingual gains in the main paper come from script-aware specialization rather than from trading off the high-resource scripts.

\subsection*{Extended Ablations}
\label{sec:supp_ablation}

The main ablation studies each MoE component in a binary on/off fashion. Here we complement it with finer-grained sweeps over the three continuous hyper-parameters that the binary study leaves open: the script-classification loss weight $\lambda_{\text{scls}}$, the shared-expert width ratio, and the router jitter $\sigma$. All runs use the identical data, schedule and evaluation protocol as the main experiment, and vary a single hyper-parameter at a time around the default configuration ($\lambda_{\text{scls}}{=}0.1$, ratio${=}0.5$, $\sigma{=}0.05$). Results are reported in Tab.~\ref{tab:supp_ablation}.

\paragraph{Script-classification loss weight.}
The main ablation shows that removing $\mathcal{L}_{\text{scls}}$ ($\lambda_{\text{scls}}{=}0$) lets the router drift away from script boundaries. The sweep in Tab.~\ref{tab:supp_ablation} shows the opposite failure mode as well: over-weighting the signal is equally harmful. A light $\lambda_{\text{scls}}{=}0.1$ is best, whereas pushing it to 0.2 or 0.5 turns the auxiliary loss into a competing objective that over-constrains the router, dropping the average to 81.44\% and 81.35\% and hitting the low-resource Tibetan hardest (down to 84.55\% at $\lambda_{\text{scls}}{=}0.5$). A gentle regularizer is therefore preferable to a hard script prior.

\paragraph{Shared-expert width ratio.}
Removing the shared expert (ratio 0.0) is the most damaging setting (81.35\% Avg), as reported before, confirming its role in preserving cross-script transfer. Among the non-zero ratios, 0.5 already recovers almost all of the benefit, and enlarging it to 1.0 does not help. 

\paragraph{Router jitter.}
A small multiplicative jitter regularizes the sample-level router. Disabling it ($\sigma{=}0$) makes routing slightly brittle (81.61\% Avg), while an overly large jitter ($\sigma{=}0.1$) injects too much noise into the routing decision and is worse (81.32\%). The default $\sigma{=}0.05$ gives the best trade-off (82.06\%).

\begin{figure}[h]
    \centering
    \includegraphics[width=0.9\linewidth]{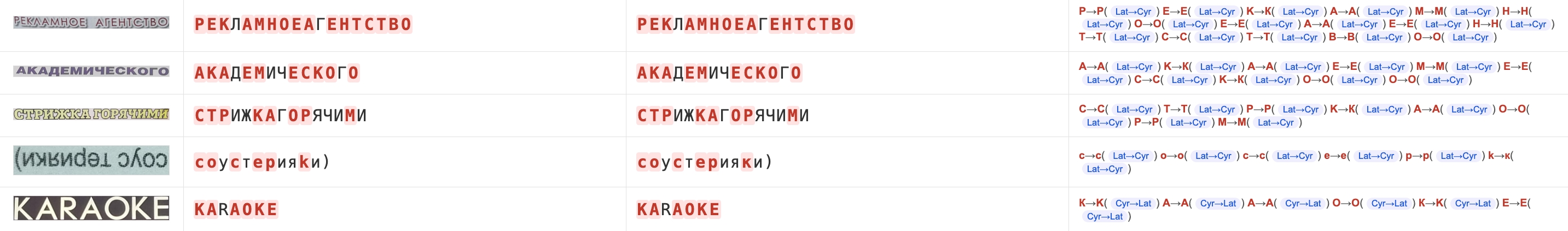}
    \caption{Russian text image, ground truth, model prediction, and homoglyph 
substitution details (left to right). Red background marks homoglyph-induced 
cross-script confusion.}
    \label{fig:lat_cyr}
\end{figure}

\subsection*{Limitations and Discussion}
\label{sec:supp_limitations}

We identify three main limitations of the current work. (1) Despite careful synthesis, a domain gap remains between our synthetic TextMuSS-10M and real-world imagery. Collecting large-scale real multilingual scene images for semi-supervised learning is a promising direction to close this gap. (2) The Latin-Cyrillic homoglyph confusion is alleviated but not fully solved. As illustrated in Fig.~\ref{fig:lat_cyr}, Cyrillic characters that are visually identical to Latin letters are occasionally mis-recognized as their Latin homoglyphs, so the prediction becomes an orthographically plausible but wrong cross-script transcription. Our fully-synthetic experiments show that a balanced data distribution effectively mitigates it, which further motivates building real Cyrillic-script data to close the remaining gap. (3) The end-to-end pipeline still depends on an upstream detector. We currently reuse the PP-OCRv5 detector, whose multilingual detection quality cannot be fully guaranteed across all scripts and may bound the overall performance.

\end{document}